# Proactive Algorithms for Job Shop Scheduling with Probabilistic Durations

**J. Christopher Beck**                                         JCB@MIE.UTORONTO.CA
*Department of Mechanical & Industrial Engineering*
*University of Toronto, Canada*

**Nic Wilson**                                                N.WILSON@4C.UCC.IE
*Cork Constraint Computation Centre*
*University College Cork, Ireland*

## Abstract

Most classical scheduling formulations assume a fixed and known duration for each activity. In this paper, we weaken this assumption, requiring instead that each duration can be represented by an independent random variable with a known mean and variance. The best solutions are ones which have a high probability of achieving a good makespan. We first create a theoretical framework, formally showing how Monte Carlo simulation can be combined with deterministic scheduling algorithms to solve this problem. We propose an associated deterministic scheduling problem whose solution is proved, under certain conditions, to be a lower bound for the probabilistic problem. We then propose and investigate a number of techniques for solving such problems based on combinations of Monte Carlo simulation, solutions to the associated deterministic problem, and either constraint programming or tabu search. Our empirical results demonstrate that a combination of the use of the associated deterministic problem and Monte Carlo simulation results in algorithms that scale best both in terms of problem size and uncertainty. Further experiments point to the correlation between the quality of the deterministic solution and the quality of the probabilistic solution as a major factor responsible for this success.

## 1. Introduction

Proactive scheduling techniques seek to produce an off-line schedule that is robust to execution time events. In this paper, we assume that we do not have perfect knowledge of the duration of each activity: the durations are determined at execution time when it is observed that an activity has finished. However, we do have partial knowledge in the form of a known probability distribution for each duration. At execution time, the activities will be dispatched according to the sequences defined by the off-line schedule and our measure of robustness is the probability with which a given quality will be achieved. More specifically, in this paper, we address the problem of job shop scheduling (and related generalizations) when the durations of the activities are random variables and the objective is to find a solution which has a high probability of having a good (ideally, minimal) makespan. This is a challenging problem as even *evaluating* a solution is a hard problem.

To address this problem, we develop a theoretical framework within which we formally define the problem and (a) construct an approach, based on Monte Carlo simulation, for evaluating both solutions and partial solutions, and (b) show that solving a carefully defined deterministic job shop scheduling problem results in a lower bound of the probabilistic minimum makespan of the probabilistic job shop scheduling problem. We use this framework to define a number of algorithms embodying three solution approaches:

1. Branch-and-bound search with Monte Carlo simulation: at each search node, the search is pruned if we can be almost certain (based on the Monte Carlo simulation) that the partial solution cannot be extended to a solution better than our current best solution.

2. Iterative deterministic search with a descending lower bound: the deterministic job shop problem whose solution is a lower bound on the probabilistic job shop problem is defined using a parameter, $q$. The lower bound proof depends on $q$ being less than or equal to $q^*(I)$, a problem-instance-dependent threshold value for problem instance $I$ that is difficult to compute. Starting with a high $q$ value, we use tree search and Monte Carlo simulation to solve a sequence of deterministic problems with decreasing $q$ values. When $q$ is large, the problems are highly constrained and easy to solve (if any solutions exist). As $q$ descends, the best probabilistic makespan from previous iterations is used to restrict the search. If we are able to reach a value of $q$ with $q \leq q^*(I)$ within the CPU time limit, then the search is approximately complete subject to the sampling error.

3. Deterministic filtering search: deterministic scheduling algorithms based on constraint programming and tabu search are used to define a number of filter-based algorithms. All these algorithms operate by generating a series of solution candidates that are evaluated by Monte Carlo simulation.

Our empirical results indicate that the Monte Carlo based branch-and-bound is only practical for very small problems. The iterative search based on descending $q$ values is as good as, or better than, the branch-and-bound algorithm on small problems, and performs significantly better on larger problems. However, even for medium-sized problems, both of these techniques are inferior to the heuristic approaches based on deterministic filtering.

**Contributions.** The main contributions of this paper are:

- the introduction of the problem of finding proactive schedules with probabilistic execution guarantees for a class of problems where the underlying deterministic scheduling problem is NP-hard;

- the development of a method for generating a lower bound on the probabilistic minimum makespan;

- the development of a particular Monte Carlo approach for evaluating solutions;

- the design and empirical analysis of a number of approximately complete and heuristic solution techniques based on either constraint-based constructive search or tabu search; and

- the identification of the correlation between deterministic and probabilistic solution quality as a key factor in the performance of the filter-based algorithms.

**Plan of Paper.** In the next section we define the probabilistic job shop scheduling problem, illustrating it with an example. Section 3 discusses related work. In Section 4, we present our theoretical framework: we formally define the problem, derive our approach for generating a lower bound based on an associated deterministic job shop problem, and show how Monte Carlo simulation can be used to evaluate solutions and partial solutions. Six search algorithms are defined in Section 5 and our empirical investigations and results appear in Section 6. In Section 7, it is shown how the results of this paper apply to much more general classes of scheduling problems. Directions for future work based on theoretical and algorithmic extensions are also discussed.

## 2. Probabilistic Job Shop Scheduling Problems

The job shop scheduling problem with probabilistic durations is a natural extension of the standard (deterministic) job shop scheduling problem (JSP).

### 2.1 Job Shop Scheduling Problems

A JSP involves a set $\mathcal{A}$ of activities, where each $A_i \in \mathcal{A}$ has a positive duration $d_i$. For each instance of a JSP, it is assumed that either all the durations are positive integers, or they are all positive real numbers.[1] $\mathcal{A}$ is partitioned into *jobs*, and each job is associated with a total ordering on that set of activities. Each activity must execute on a specified unary capacity resource. No activities that require the same resource can overlap in their execution, and once an activity is started it must be executed for its entire duration. We represent this formally by another partition of $\mathcal{A}$ into *resource sets*: two activities are in the same resource set if and only if they require the same resource.

A *solution* consists of a total ordering on each resource set, which does not conflict with the jobs ordering, i.e., the union of the resource orderings and job orderings is an acyclic relation on $\mathcal{A}$. Thus, if $A_i$ and $A_j$ are in the same resource set, a solution either orders $A_i$ before $A_j$ (meaning that $A_j$ starts no sooner than the end of $A_i$), or $A_j$ before $A_i$. The set of solutions of a job shop problem will be labeled $S$. A *partial solution* consists of a partial ordering on each resource set which can be extended to a solution.

Let $s$ be a (partial) solution. A *path in $s$* (or an *s-path*) is a sequence of activities such that if $A_i$ immediately precedes $A_j$ in the sequence, then either (i) $A_i$ and $A_j$ are in the same job, and $A_i$ precedes $A_j$ in that job, or (ii) $A_i$ and $A_j$ are in the same resource set and $s$ orders $A_i$ before $A_j$. The length, $len(\pi)$, of a path $\pi$ (of a solution) is equal to the sum of the durations of the activities in the path, i.e., $\sum_{A_i \in \pi} d_i$. The *makespan*, $make(s)$, of a solution $s$ is defined to be the length of a longest $s$-path. An $s$-path, $\pi$, is said to be a *critical s-path* if the length of $\pi$ is equal to the makespan of the solution $s$, i.e., it is one of the longest $s$-paths. The *minimum makespan* of a job shop scheduling problem is defined to be the minimum value of $make(s)$ over all solutions $s$.

The above definitions focus on solutions rather than on schedules. Here, we briefly indicate how our definitions relate to, perhaps more immediately intuitive, definitions focusing on schedules. A *schedule* assigns the start time of each activity, and so can be considered as a function from the set of activities $\mathcal{A}$ to the set of time-points, defining when each activity starts. The set of time-points is assumed to be either the set of non-negative integers or the set of non-negative real numbers. Let $start_i$ be the start time of activity $A_i \in \mathcal{A}$ with respect to a particular schedule, and let $end_i$, its end time, be $start_i + d_i$. For $A_i, A_j \in \mathcal{A}$, write $A_i \prec A_j$ for the constraint $end_i \leq start_j$. A schedule is defined to be valid if the following two conditions hold for any two different activities $A_i, A_j \in \mathcal{A}$: (a) if $A_i$ precedes $A_j$ in the same job, then $A_i \prec A_j$; and (b) if $A_i$ and $A_j$ are in the same resource set, then either $A_i \prec A_j$ or $A_j \prec A_i$ (since $A_i$ and $A_j$ are not allowed to overlap).

---

1. Our empirical investigations examine the integer case. As shown below, the theoretical results hold also for the case of positive real number durations.

Let $Z$ be a valid schedule. Define $make(Z)$, the makespan of $Z$, to be $\max_{A_i \in \mathcal{A}} end_i$, the time at which the last activity has been completed. The minimum makespan is defined to be the minimum value of $make(Z)$ over all valid schedules.

Each solution $s$ defines a valid schedule $sched(s)$, where each activity is started as soon as its immediate predecessors (if any) have finished, and activities without predecessors are started at time-point 0 (so $sched(s)$ is a non-delay schedule given the precedence constraints expressed by $s$). An immediate predecessor of activity $A_j$ with respect to a particular solution is defined to be an activity which is an immediate predecessor of $A_j$ either with respect to the ordering on the job containing $A_j$, or with respect to the ordering (associated with the solution) on the resource set containing $A_j$. It can be shown that the makespan of $sched(s)$ is equal to $make(s)$ as defined earlier, hence justifying our definition.

Conversely, given a valid schedule $Z$, we can define a solution, which we call $sol(Z)$, by ordering each resource set with the relation $\prec$ defined above. If $Z$ is a schedule, then the makespan of $sched(sol(Z))$, which is equal to $make(sol(Z))$, is less than or equal to the makespan of $Z$. This implies that the minimum makespan over all solutions is equal to the minimum makespan over all valid schedules. Therefore, if we are interested in schedules with the best makespans, we need only consider solutions and their associated schedules.

To summarize, when aiming to find the minimum makespan for a JSP, we can focus on searching over all solutions, rather than over all schedules, because (i) for any schedule $Z$, there exists a solution $s = sol(Z)$ such that $Z$ is consistent with $s$ (i.e., satisfies the precedence constraints expressed by $s$); and (ii) for any solution $s$, we can efficiently construct a schedule $sched(s)$ which is optimal among schedules consistent with $s$ (and furthermore, the makespan of $sched(s)$ is equal to $make(s)$).

**JSP Example.** Consider a job shop scheduling problem involving two jobs and five activities as shown in Figure 1. The first job consists of the sequence $(A_1, A_2, A_3)$ of activities; the second job consists of the sequence $(A_4, A_5)$. There are three resources involved. $A_1$ and $A_4$ require the first resource; hence activities $A_1$ and $A_4$ cannot overlap, and so either (i) $A_1$ precedes $A_4$, or (ii) $A_4$ precedes $A_1$. Activities $A_3$ and $A_5$ require the second resource; $A_2$ requires the third resource. Hence, the resource sets are $\{A_1, A_4\}$, $\{A_2\}$ and $\{A_3, A_5\}$. There are four solutions:

- $s_a$ involves the orderings $A_1 \prec A_4$ and $A_3 \prec A_5$;

- $s_b$ is defined by $A_1 \prec A_4$ and $A_5 \prec A_3$;

- $s_c$ by $A_4 \prec A_1$ and $A_3 \prec A_5$; and

- $s_d$ by $A_4 \prec A_1$ and $A_5 \prec A_3$.

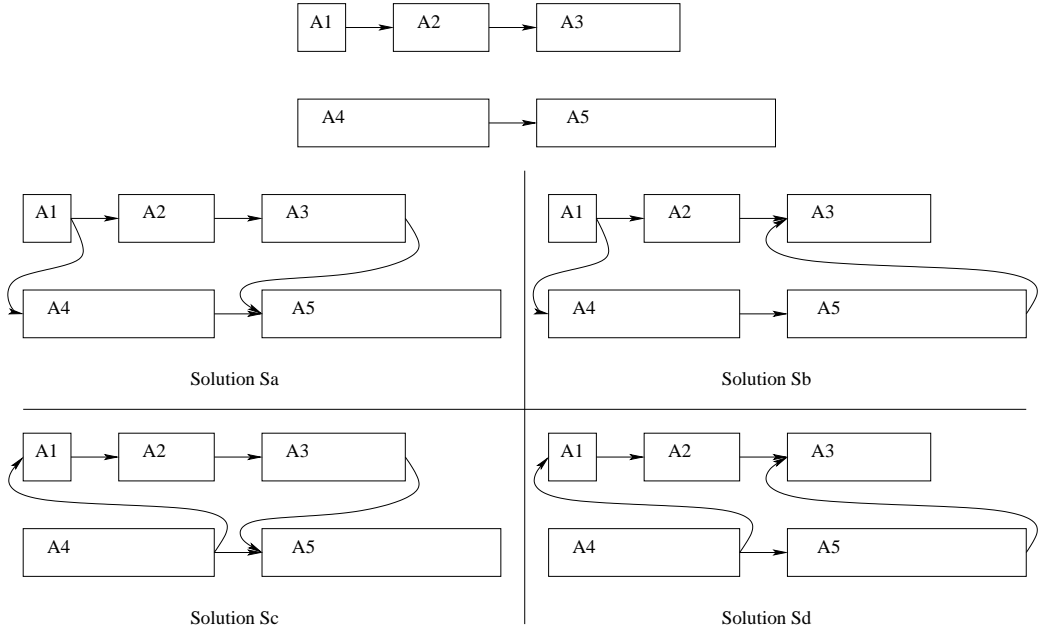

Figure 1: The example JSP with its four solutions.

The duration of activity $A_i$ is $d_i$. The sequence $(A_1, A_4, A_5)$ is an $s_a$-path, whose length is $d_1 + d_4 + d_5$. Also, if $\pi$ is the $s_a$-path $(A_1, A_2, A_3, A_5)$, then $len(\pi) = d_1 + d_2 + d_3 + d_5$. The only other $s_a$-paths are subsequences of one of these two. Hence, $make(s_a)$, the makespan of solution $s_a$, is equal to $\max(d_1 + d_4 + d_5, d_1 + d_2 + d_3 + d_5) = d_1 + d_5 + \max(d_4, d_2 + d_3)$. In particular, if $d_1 = 1, d_2 = 2, d_3 = 3, d_4 = 4$ and $d_5 = 5$, then $make(s_a) = 11$ time units. We then also have $make(s_b) = 13$, $make(s_c) = 15$ and $make(s_d) = 12$. Hence, the minimum makespan is $make(s_a) = 11$.

Let $Z = sched(s_a)$ be the schedule associated with solution $s_a$. This is generated as follows. $A_1$ has no predecessors, so we start $A_1$ at the beginning, setting $Z(A_1) = 0$; hence activity $A_1$ starts at time-point 0 and ends at time-point $d_1$. The only predecessor of $A_4$ is $A_1$, so we set $Z(A_4) = d_1$. Similarly, we set $Z(A_2) = d_1$, and so activity $A_2$ ends at time-point $d_1 + d_2$. Continuing, we set $Z(A_3) = d_1 + d_2$. Activity $A_5$ has two immediate predecessors (for this solution, $s_a$), $A_3$ and $A_4$, and so $A_5$ is set to start as soon as both of these activities have been completed, which is at time-point $\max(d_1 + d_2 + d_3, d_1 + d_4)$. All activities have been completed when $A_5$ has been completed, which is at time-point $\max(d_1 + d_2 + d_3, d_1 + d_4) + d_5 = d_1 + d_5 + \max(d_4, d_2 + d_3)$. This confirms that the makespan $make(s_a)$ of solution $s_a$ is equal to the makespan of its associated schedule $sched(s_a)$.

## 2.2 Independent and General Probabilistic Job Shop Scheduling Problems

An *independent probabilistic job shop scheduling problem* is defined in the same way as a JSP, except that the duration $\mathbf{d}_i$ associated with an activity $A_i \in \mathcal{A}$ is a random variable; we assume that in each instance of a probabilistic JSP, either all the durations are positive integer-valued random variables, or they are all positive real-valued random variables. $\mathbf{d}_i$ has (known) distribution $P_i$, expected value $\mu_i = \mathrm{E}[\mathbf{d_i}]$ and variance $\sigma_i^2 = \mathrm{Var}[\mathbf{d_i}]$. These random variables are fully independent. The length of a path $\pi$ of a solution $s$ is now a random variable, which we write as $\mathbf{len}(\pi)$. The makespan $\mathbf{make}(s)$ of solution $s$ (the length of the longest path in $s$) is therefore also a random variable, which we will sometimes refer to as the *random makespan* of $s$.

We can generalize this to the non-independent case. In the *probabilistic job shop scheduling problem* we have a joint probability measure $P$ over the durations vectors. (The intention is that we can efficiently sample with the joint density function. For example, a Bayesian network might be used to represent $P$.) Here, for activity $A_i$, distribution $P_i$ is defined to be the appropriate marginal distribution, with expected value $\mu_i$ and variance $\sigma_i^2$.

Loosely speaking, in a probabilistic job shop scheduling problem, we want to find as small a value of $D$ as possible such that there is a solution whose random makespan is, with high probability, less than $D$ (the "deadline" for all activities to finish). This time value $D$ will be called the probabilistic minimum makespan.

Evaluating a solution for a deterministic JSP, i.e., finding the associated makespan given a duration for each activity, can be achieved in low degree polynomial time using a longest path algorithm. Without the ordering on each resource set, the disjunctions of resource constraints that must be satisfied to find a solution turn this very easy problem into the NP-complete JSP (Garey & Johnson, 1979). PERT networks, on the other hand, generalize the simple longest-path problem by allowing durations to be independent random variables, leading to a #P-complete problem (Hagstrom, 1988). The probabilistic JSP makes both these generalizations. Consequently, finding the optimal solutions of a probabilistic JSP appears to be very hard, and we focus on methods for finding good solutions instead.

Evaluating (approximately) a solution of a probabilistic JSP can be done relatively efficiently using Monte Carlo simulation: for each of a large number of trials we randomly sample the duration of every activity and generate the makespan associated with that trial. Roughly speaking, we approximately evaluate the solution by evaluating the sampled distribution of these makespans. This approach is described in detail in Section 4.3.

Almost all of our solution techniques involve associating a deterministic job shop problem with the given probabilistic job shop problem, by replacing, for some number $q$, each random duration by the mean of its distribution plus $q$ times its standard deviation. Hence, we set the duration $d_i$ of activity $A_i$ in the associated deterministic problem to be $\mu_i + q \times \sigma_i$ for the case of continuous time. For the case when time-points are integers, we set $d_i = \lfloor \mu_i + q \times \sigma_i \rfloor$. For certain values of $q$, this leads to the minimum makespan of the deterministic problem being a lower bound for the probabilistic minimum makespan, as shown in Section 4.2. This lower bound can be useful for pruning in a branch-and-bound algorithm. More generally, we show how solving the associated deterministic problem can be used to help solve the probabilistic problem.

Our assumptions about the joint probability are somewhat restrictive. For example, the model does not allow an activity's duration to depend on its start time; however, it can be extended to certain situations of this kind.[2] Despite these restrictions (which are common in related literature—see Section 3), our model does apply to an interesting class of problems

---

2. We could allow the duration of each activity to be probabilistically dependent only on its start time, given the additional (very natural) coherence condition that for any time-point $t'$, the conditional probability that $end_i \geq t'$, given $start_i = t$, is monotonically increasing in $t$, i.e., $\Pr(end_i \geq t' | start_i = t_1) \leq \Pr(end_i \geq t' | start_i = t_2)$ if $t_1 \leq t_2$. This condition ensures that, for any given solution, there is no

that has not been previously addressed. Extending our model to richer representations by relaxing our assumptions remains for future work.

**Probabilistic JSP Example.** We consider an independent probabilistic job shop scheduling problem with the same structure as the JSP example in Figure 1. The durations of activities $A_2$, $A_3$ and $A_4$ are now independent real-valued random variables (referred to as $\mathbf{d}_2$, $\mathbf{d}_3$ and $\mathbf{d}_4$, respectively) which are all approximately normally distributed with standard deviation 0.5 ($\sigma_2 = \sigma_3 = \sigma_4 = 0.5$) and with means $\mu_2 = 2$, $\mu_3 = 3$ and $\mu_4 = 4$. The durations of activities $A_1$ and $A_5$ are deterministic, being equal to 1 and 5, respectively.

Let $\pi$ be the $s_a$-path $(A_1, A_2, A_3, A_5)$. The length $\mathbf{len}(\pi)$ of $\pi$ is an approximately normally distributed random variable with mean $1+2+3+5 = 11$ and variance $0.5^2+0.5^2 = 0.5$ and hence standard deviation $1/\sqrt{2}$.

The length of $s_a$-path $\pi' = (A_1, A_4, A_5)$ is an approximately normal random variable with mean 10 and standard deviation 0.5. The (random) makespan $\mathbf{make}(s_a)$ of solution $s_a$ is a random variable equaling the maximum of random variables $\mathbf{len}(\pi)$ and $\mathbf{len}(\pi')$. In general, the maximum of two independent normally distributed random variables is not normally distributed; however, $\pi$ is, with high probability, longer than $\pi'$, so the distribution of $\mathbf{make}(s_a)$ is approximately equal to the distribution of $\mathbf{len}(\pi)$.

## 3. Previous Work

There has been considerable work on scheduling with uncertainty in a variety of fields including artificial intelligence (AI), operations research (OR), fault-tolerant computing, and systems. For surveys of the literature, mostly focusing in AI and OR, see the work of Davenport and Beck (2000), Herroelen and Leus (2005), and Bidot (2005).

At the highest level, there are two approaches to such problems: *proactive* scheduling, where some knowledge of the uncertainty is taken into account when generating an off-line schedule; and *reactive* scheduling where decisions are made on-line to deal with unexpected changes. While there is significant work in reactive scheduling and, indeed, on techniques that combine reactive and proactive scheduling such as least commitment approaches (see the surveys noted above), here our interest is on pure proactive scheduling. Three categories of proactive approaches have been identified: redundancy-based techniques, probabilistic techniques, and contingent/policy-based techniques (Herroelen & Leus, 2005). We briefly look at each of these in turn.

### 3.1 Redundancy-based Techniques

Redundancy-based techniques generate a schedule that includes the allocation of extra resources and/or time in the schedule. The intuition is that these redundant allocations will help to cushion the impact of unexpected events during execution. For example, extra time can be "consumed" when an activity takes longer than expected to execute. Because there is a clear conflict between insertion of redundancy and common measures of schedule quality (e.g., makespan), the focus of the work tends to be the intelligent insertion of redundancy in order to achieve a satisfactory trade-off between schedule quality and robustness.

---

advantage in delaying starting an activity when its predecessors have finished. Allowing such a delay would break the assumptions underlying our formulation.

It is common in fault-tolerant scheduling with real-time guarantees to reserve redundant resources (i.e., processors) or time. In the former case, multiple instantiations of a given process are executed in parallel and error detection can be done by comparing the results of the different instantiations. In contrast, in time redundancy, some time is reserved for re-execution of a process that fails. Given a fault model, either technique can be used to provide real-time guarantees (Ghosh, Melhem, & Mossé, 1995; Ghosh, 1996).

A similar approach is used in the work of Gao (1995) and Davenport, Gefflot and Beck (2001) in the context of job shop scheduling. Statistical information about the mean time between failure and the mean repair time of machines is used to either extend the duration of critical activities in the former work or to require that any solution produced must respect constraints on the slack of each activity. Given a solution, the slack is the room that an activity has to "move" without breaking a constraint or increasing the cost. Typically, it is formalized as the difference between an activity's possible time window in a solution (i.e., its latest possible end time less its earliest possible start time) and the duration of the activity. The advantage of Gao's approach is that it is purely a modeling approach: the problem is changed to incorporate extended durations and any scheduling techniques can be used to solve the problem. However, Davenport et al. show that reasoning about the slack shared amongst a set of activities can lead to better solutions at the cost of specialized solving approaches.

Leon, Wu and Storer (1994) present an approach to job shop scheduling where the objective function is modified to be a linear combination of the expected makespan and expected delay assuming that machines can break down and that, at execution time, disruptions are dealt with by shifting activities later in time while maintaining the sequence in the original schedule. While this basic technique is more properly seen as a probabilistic approach, the authors show that an exact calculation of this measure is intractable unless a single disruption is assumed. When there are likely to be multiple disruptions, the authors present a number of surrogate measures. Empirically, the best surrogate measure is the deterministic makespan minus the mean activity slack. Unlike, Gao and Davenport et al., Leon et al. provide a more formal probabilistic foundation, but temporal redundancy plays a central role in the practical application of their approach.

## 3.2 Probabilistic Techniques

Probabilistic techniques use representations of uncertainty to reason about likely outcomes when the schedule is executed.[3] Rather than explicitly inserting redundancy in an attempt to create a robust schedule, probabilistic techniques build a schedule that optimizes some measure of probabilistic performance. Performance measures typically come in two forms: an expected value such as expected makespan or expected weighted tardiness, and a probabilistic guarantee with respect to a threshold value of a deterministic optimization measure. An example of the latter measure, as discussed below, is the probability that the flow time of a schedule will be less than a particular value.

Optimal expected value scheduling problems have been widely studied in OR (Pinedo, 2003). In many cases, the approach takes the form of dispatch rules or slightly more complicated polynomial time algorithms that will find the optimal schedule for tractable problems (e.g., 1 and 2 machine problems) and which serve as heuristics for more difficult problems. One example of such work in the AI literature is that of Wurman and Wellman (1996) which extends decision theoretic planning concepts to scheduling. The problem studied assumes a single machine, stochastic processing time and stochastic set-up time, and has as its objective the minimization of the expected weighted number of tardy jobs. The authors propose a state-space search and solve the problem of multi-objective stochastic dominance A*. Critical aspects of this work are the use of a number of sophisticated path pruning rules and relaxation-based heuristics for the evaluation of promising nodes.

---

3. Alternative representations of uncertainty such as fuzzy sets can also be used (Herroelen & Leus, 2005).

A threshold measure is used by Burns, Punnekkat, Littlewood and Wright (1997) in a fault-tolerant, single processor, pre-emptive scheduling application. The objective is to find the minimum fault arrival rate such that all tasks can be scheduled to meet their deadlines. Based on a fault-model, the probability of observing that fault arrival rate is calculated and used as a measure of schedule quality. The optimization problem, then, is to find the schedule that maximizes the probability of all tasks meeting their deadlines under the fault arrival process.

In a one-machine manufacturing context with independent activities, Daniels and Carrillo (1997) define a $\beta$-robust schedule as the sequence that maximizes the probability that the execution will achieve a flow time no greater than a given threshold. While the underlying deterministic scheduling problem is solvable in polynomial time and, indeed, the minimum expected flow time schedule can be found in polynomial time, it is shown that finding the $\beta$-robust schedule is NP-hard. Daniels and Carrillo present branch-and-bound and heuristic techniques to solve this problem.

## 3.3 Contingent and Policy-based Approaches

Unlike the approaches described above, contingent and policy-based approaches do not generate a single off-line schedule. Rather, what is produced is a branching or contingent schedule or, in the extreme, a policy, that specifies the actions to be taken when a particular set of circumstances arises. Given the importance of having an off-line schedule in terms of coordination with other entities in the context surrounding the scheduling problem, this difference can have significant practical implications (see Herroelen & Leus, 2005, for a discussion).

An elegant example of a contingent scheduling approach is the "just-in-case" work of Drummond, Bresina and Swanson (1994). Given an initial, deterministic schedule for a single telescope observation problem, the approach identifies the activity most likely to fail based on the available uncertainty information. At this point, a new schedule is produced assuming the activity does, indeed, fail. Repeated application of the identification of the most-likely-to-fail activity and generation of a new schedule results in a branching schedule where a number of the most likely contingencies are accounted for in alternative schedules. At execution time, when an activity fails, the execution switches to the alternative schedule if one exists. If an alternative does not exist, on-line rescheduling is done. Empirical results demonstrate that a significantly larger portion of an existing (branching) schedule can be executed without having to revert to rescheduling as compared to the original deterministic schedule.

One of the weaknesses of the just-in-case scheduling surrounds the combinatorics of multiple resources. With multiple inter-dependent telescopes, the problem quickly becomes intractable. Policy-based approaches such as Markov Decision Processes (MDPs) (Boutilier, Dean, & Hanks, 1999) have been applied to such problems. Here, the objective is to produce a policy mapping states to actions that will direct the on-line execution of the schedule: when a given state is encountered, the corresponding action is taken. Meuleau et al. (1998) apply MDPs to a stochastic military resource allocation problem where weapons must be allocated to targets. Given a limited number of weapons and uncertainty about the effectiveness of a given allocation, an MDP is used to derive an optimal policy where the states are represented by the number of remaining weapons and targets, and the actions are the weapon allocation decisions. The goal is to minimize the expected number of surviving targets. Empirical results demonstrated the computational challenges of such an approach as a 6 target, 60 weapon problem required approximately 6 hours of CPU time (albeit on now-outdated hardware).

In the OR literature, there has been substantial work (cited in Brucker, Drexl, Möhring, Neumann and Pesch, 1999, and Herroelen and Leus, 2005) on stochastic resource-constraint project scheduling, a generalization of job shop scheduling. The general form of these approaches is a multi-stage stochastic programming problem, with the objective of finding a scheduling policy which will minimize the expected makespan. In this context, a scheduling policy makes decisions on-line about what activities to execute. Decisions need to be made at the beginning of the schedule and at the end time of each activity, and the information used for such decisions need be only that which has become known before the time of decision making. A number of different classes of policy have been investigated. For example, a *minimal forbidden subset* of activities, $F$, is a set such that the activities in $F$ cannot be executed simultaneously due to resource constraints, but that any subset of $F$ can be so executed. A *pre-selective* policy identifies such a set $F$ and a *waiting activity*, $j \in F$, such that $j$ cannot be started until at least one activity $i \in F - \{j\}$ has been executed. During execution, $j$ can be started only when at least one other activity in $F$ has finished. The proactive problem, then, is to identify the waiting activity for each minimal forbidden subset such that the expected makespan is minimized. The computational challenges of pre-selective policies (in particular, due to the number of minimal forbidden subsets) have led to work on different classes of policy as well as heuristic approaches.

## 3.4 Discussion

The work in this paper falls within the probabilistic scheduling approaches and is most closely inspired by the $\beta$-robustness work of Daniels and Carrillo (1997). However, unlike Daniels and Carrillo, we address a scheduling model where the deterministic problem that underlies the probabilistic job shop scheduling problem is, itself, NP-hard. This is the first work of which we are aware that seeks to provide probabilistic guarantees where the underlying deterministic problem is computationally difficult.

## 4. Theoretical Framework

In this section, we develop our theoretical framework for probabilistic job shop problems. In Section 4.1, we define how we compare solutions, using what we call $\alpha$-makespans. If the $\alpha$-makespan of solution $s$ is less than time value $D$, then there is at least chance $1 - \alpha$ that the (random) makespan of $s$ is less than $D$. As it can be useful to have an idea about how far a solution's $\alpha$-makespan is from the optimum $\alpha$-makespan (i.e., the minimum $\alpha$-makespan over all solutions), in Section 4.2, we describe an approach for finding a lower bound for the optimum $\alpha$-makespan. Section 4.3 considers the problem of evaluating a given solution, $s$, by using Monte Carlo simulation to estimate the $\alpha$-makespan of $s$.

In order to separate our theoretical contributions from our empirical analysis, we summarize the notation introduced in this section in Section 5.1. Readers interested primarily in the algorithms and empirical results can therefore move directly to Section 5.

This section makes use of notation introduced in Section 2: the definitions in Section 2.1 of a JSP, a solution, paths in a solution, the makespan of a solution, and the minimum makespan; and the definitions in Section 2.2 of a probabilistic JSP and the random makespan of a solution.

## 4.1 Comparing Solutions and Probabilistic Makespan

In a standard job shop problem, solutions can be compared by considering the associated makespans. In the probabilistic case, the makespan of a solution is a random variable, so comparing solutions is less straight-forward. We map the random makespan to a scalar quantity, called the $\alpha$-makespan, which sums up how good it is; solutions are compared by comparing their associated $\alpha$-makespans. A simple idea is to prefer solutions with smaller expected makespan. However, there may be a substantial probability that the makespan of the solution will be much higher than its expected value. Instead, we take the following approach: if we can be confident that the random makespan for solution $s$ is at most $D$, but we cannot be confident that the makespan for solution $s'$ is at most $D$, then we prefer solution $s$ to solution $s'$.

We fix a value $\alpha$, which is used to bound probabilities. Although we imagine that in most natural applications of this work, $\alpha$ would be quite small (e.g., less than 0.1) we assume only that $\alpha$ is in the range $(0, 0.5]$. If the probability of an event is at least $1 - \alpha$, then we say that the event is *sufficiently certain*. The experiments described in Section 6 use a value of $\alpha = 0.05$, so that "sufficiently certain" then means "occurs with at least 95% chance".

Let $D$ be a time value, and let $s$ be a solution. $D$ is said to be $\alpha$-*achievable using* $s$ if it is sufficiently certain that all jobs finish by $D$ when we use solution $s$; that is, if $\Pr(\mathbf{make}(s) \leq D) \geq 1 - \alpha$, where $\mathbf{make}(s)$ is the random makespan of $s$.

$D$ is said to be $\alpha$-*achievable* if there is some solution $s$ such that $D$ is $\alpha$-achievable using $s$, i.e., if there exists some solution $s$ making it sufficiently certain that all jobs finish by $D$. Time value $D$ is $\alpha$-achievable if and only if $\max_{s \in S} \Pr(\mathbf{make}(s) \leq D)) \geq 1 - \alpha$, where the max is over all solutions $s$.

Define $Ach_\alpha(s)$ to be the set of all $D$ which are $\alpha$-achievable using $s$. We define $D_\alpha(s)$, *the $\alpha$-makespan of $s$*, to be the infimum[4] of $Ach_\alpha(s)$. Then $D_\alpha$, *the $\alpha$-minimum makespan*, is defined to be the infimum of $Ach_\alpha$, which is the set of all $D$ which are $\alpha$-achievable, so $D_\alpha = \inf \{D : (\max_{s \in S} \Pr(\mathbf{make}(s) \le D)) \ge 1 - \alpha\}$. We will also sometimes refer to $D_\alpha(s)$ as the *probabilistic makespan* of $s$, and refer to $D_\alpha$ as the *probabilistic minimum makespan*.[5] We prefer solutions which have better (i.e., smaller) $\alpha$-makespans. Equivalently, solution $s$ is considered better than $s'$ if there is a time value $D$ which is $\alpha$-achievable using $s$ but not $\alpha$-achievable using $s'$. Optimal solutions are ones whose $\alpha$-makespan is equal to the $\alpha$-minimum makespan.

---

4. That is, the greatest lower bound of $Ach_\alpha(s)$; in fact, as shown by Proposition 1(i), $D_\alpha(s)$ is the smallest element of $Ach_\alpha(s)$. Hence, $Ach_\alpha(s)$ is equal to the closed interval $[D_\alpha(s), \infty)$, i.e., the set of time-points $D$ such that $D \geq D_\alpha(s)$.

We prove some technical properties of $\alpha$-makespans and $\alpha$-achievability relevant for mathematical results in later sections. In particular, Proposition 1(ii) states that the $\alpha$-minimum makespan $D_\alpha$ is $\alpha$-achievable: i.e., there exists some solution which makes it sufficiently certain that all jobs finish by $D_\alpha$. $D_\alpha$ is the smallest value satisfying this property.

**Lemma 1** *With the above notation:*

(i) $Ach_\alpha = \bigcup_{s \in S} Ach_\alpha(s)$;

(ii) *there exists a solution $s$ such that $Ach_\alpha = Ach_\alpha(s)$ and $D_\alpha = D_\alpha(s)$;*

(iii) $D_\alpha = \min_{s \in S} D_\alpha(s)$, *the minimum of $D_\alpha(s)$ over all solutions $s$.*

*Proof:*

(i) $D$ is $\alpha$-achievable if and only if for some solution $s$, $D \in Ach_\alpha(s)$, which is true if and only if $D \in \bigcup_{s \in S} Ach_\alpha(s)$.

(ii) Consider the following property $(*)$ on set of time values $A$: if $D \in A$ and $D'$ is a time value greater than $D$ (i.e., $D' > D$), then $D' \in A$; that is, $A$ is an interval with no upper bound. Let $A$ and $B$ be two sets with property $(*)$; then either $A \subseteq B$ or $B \subseteq A$. (To show this, suppose otherwise, that neither $A \subseteq B$ nor $B \subseteq A$; then there exists some $x \in A - B$ and some $y \in B - A$; $x$ and $y$ must be different, and so we can assume, without loss of generality, that $x < y$; then by property $(*)$, $y \in A$ which is the contradiction required.) Hence, $A \cup B$ is either equal to $A$ or equal to $B$. By using induction, it follows that the union of a finite number of sets with property $(*)$ is one of the sets. Each set $Ach_\alpha(s)$ satisfies property $(*)$; therefore, $\bigcup_{s \in S} Ach_\alpha(s) = Ach_\alpha^{s_0}$ for some solution $s_0$, so, by (i), $Ach_\alpha = Ach_\alpha^{s_0}$. This implies also $D_\alpha = D_\alpha(s_0)$.

(iii) Let $s$ be any solution and let $D$ be any time value. Clearly, if $D$ is $\alpha$-achievable using $s$, then $D$ is $\alpha$-achievable. This implies that $D_\alpha \le D_\alpha(s)$. Hence, $D_\alpha \le \min_{s \in S} D_\alpha(s)$. By (ii), $D_\alpha = D_\alpha(s)$ for some solution $s$, so $D_\alpha = \min_{s \in S} D_\alpha(s)$, as required. $\square$

**Proposition 1**

(i) *Let $s$ be any solution. $D_\alpha(s)$ is $\alpha$-achievable using $s$, i.e., $\Pr(\mathbf{make}(s) \le D_\alpha(s)) \ge 1 - \alpha$.*

(ii) $D_\alpha$ *is $\alpha$-achievable, i.e., there exists some solution $s$ with $\Pr(\mathbf{make}(s) \le D_\alpha) \ge 1 - \alpha$.*

---

5. Note that the probabilistic makespan is a number (a time value), as opposed to the random makespan of a solution, which is a random variable.

*Proof:*

In the discrete case, when the set of time values is the set of non-negative integers, then the infimum in the definitions of $D_\alpha(s)$ and $D_\alpha$ is the same as minimum. (i) and (ii) then follow immediately from the definitions.

We now consider the case when the set of time values is the set of non-negative real numbers.

(i): For $m, n \in \{1, 2, \ldots, \}$, let $G_m = \Pr(0 < \mathbf{make}(s) - D_\alpha(s) \leq \frac{1}{m})$, and let $g_n = \Pr(\frac{1}{n+1} < \mathbf{make}(s) - D_\alpha(s) \leq \frac{1}{n})$. By the countable additivity axiom of probability measures, $G_m = \sum_{n=m}^{\infty} g_n$. This means that $\sum_{n=m}^{l-1} g_n$ tends to $G_m$ as $l$ tends to infinity, and hence $G_l = \sum_{l}^{\infty} g_n = G_m - \sum_{n=m}^{l-1} g_n$ tends to 0. So, we have $\lim_{m \to \infty} G_m = 0$. For all $m > 0$, we have $\Pr(\mathbf{make}(s) \leq D_\alpha(s) + \frac{1}{m}) \geq 1 - \alpha$, by definition of $D_\alpha(s)$. Also $\Pr(\mathbf{make}(s) \leq D_\alpha(s)) + G_m = \Pr(\mathbf{make}(s) \leq D_\alpha(s) + \frac{1}{m})$. So, for all $m = 1, 2, \ldots$, $\Pr(\mathbf{make}(s) \leq D_\alpha(s)) \geq 1 - \alpha - G_m$, which implies $\Pr(\mathbf{make}(s) \leq D_\alpha(s)) \geq 1 - \alpha$, because $G_m$ tends to 0 as $m$ tends to infinity.

(ii): By part (ii) of Lemma 1, for some solution $s$, $D_\alpha = D_\alpha(s)$. Part (i) then implies that $\Pr(\mathbf{make}(s) \leq D_\alpha) \geq 1 - \alpha$. □

**Probabilistic JSP Example continued.** We continue the example from Section 2.1 and Section 2.2. Set $\alpha$ to 0.05, corresponding to 95% confidence. A value of $D = 12.5$ is $\alpha$-achievable using solution $s_a$, since there is more than 95% chance that both paths $\pi$ and $\pi'$ are (simultaneously) shorter than length 12.5, and so the probability that the random makespan $\mathbf{make}(s_a)$ is less than 12.5 is more than 0.95.

Now consider a value of $D = 12.0$. Since $\mathbf{len}(\pi)$ (the random length of $\pi$) has mean 11 and standard deviation $1/\sqrt{2}$, the chance that $\mathbf{len}(\pi) \leq 12.0$ is approximately the chance that a normal distribution is no more than $\sqrt{2}$ standard deviations above its mean; this probability is about 0.92. Therefore, $D = 12.0$ is not $\alpha$-achievable using solution $s_a$, since there is less than 0.95 chance that the random makespan $\mathbf{make}(s_a)$ is no more than $D$.

The $\alpha$-makespan (also referred to as the "probabilistic makespan") of solution $s_a$ is therefore between 12.0 and 12.5. In fact, the $\alpha$-makespan $D_\alpha(s_a)$ is approximately equal to 12.16, since there is approximately 95% chance that the (random) makespan $\mathbf{make}(s_a)$ is at most 12.16. It is easy to show that $D = 12.16$ is not $\alpha$-achievable using any other solution, so $D_\alpha$, the $\alpha$-minimum makespan, is equal to $D_\alpha(s_a)$, and hence about 12.16.

## 4.2 A Lower Bound For $\alpha$-Minimum Makespan

In this section we show that a lower bound for the $\alpha$-minimum makespan $D_\alpha$ can be found by solving a particular deterministic JSP.

A common approach is to generate a deterministic problem by replacing each random duration by the mean of the distribution. As we show, under certain conditions, the minimum makespan of this deterministic JSP is a lower bound for the probabilistic minimum makespan. For instance, in the example, the minimum makespan of such a deterministic JSP is 11, and the probabilistic minimum makespan is about 12.16. However, an obvious weakness with this approach is that it does not take into account the spreads of the distributions. This is especially important since we are typically considering a small value of $\alpha$, such as 0.05. We can generate a stronger lower bound by taking into account the variances of the distributions when generating the associated deterministic job shop problem.

**Generating a Deterministic JSP from a Probabilistic JSP and a Value $q$.** From a probabilistic job shop problem, we will generate a particular deterministic job shop problem, depending on a parameter $q \geq 0$. We will use this transformation for almost all the algorithms in Section 5. The deterministic JSP is the same as the probabilistic JSP except with each random duration replaced by a particular time value. Solving the corresponding deterministic problem will give us information about the probabilistic problem. The deterministic JSP consists of the same set $\mathcal{A}$ of activities, partitioned into the same resource sets and the same jobs, with the same total order on each job. The duration of activity $A_i$ in the deterministic problem is defined to be $\mu_i + q\sigma_i$, where $\mu_i$ and $\sigma_i$ are respectively the mean and standard deviation of the duration of activity $A_i$ in the probabilistic job shop problem. Hence, if $q = 0$, the associated deterministic problem corresponds to replacing each random duration by its mean. Let $make_q(s)$ be the deterministic makespan of solution $s$, i.e., the makespan of $s$ for the associated deterministic problem (which is defined to be the length of the longest $s$-path—see Section 2.1). Let $make_q$ be the minimum deterministic makespan over all solutions.

Let $s$ be a solution. We say that $s$ is *probabilistically optimal* if $D_\alpha(s) = D_\alpha$. Let $\pi$ be an $s$-path. ($\pi$ is a path in both the probabilistic and deterministic problems.) $\pi$ is said to be a (deterministically) critical path if it is a critical path in the deterministic problem. The length of $\pi$ in the deterministic problem, $len_q(\pi)$, is equal to the sum of the durations of activities in the path: $\sum_{A_i \in \pi}(\mu_i + q\sigma_i)$, which equals $\sum_{A_i \in \pi} \mu_i + q \sum_{A_i \in \pi} \sigma_i$.

We introduce the following rather technical definition whose significance is made clear by Proposition 2: $q$ is $\alpha$-*sufficient* if there exists a (deterministically) critical path $\pi$ in some probabilistically optimal solution $s$ with $\Pr(\mathbf{len}(\pi) > len_q(\pi)) > \alpha$, i.e., there is more than $\alpha$ chance that the random path length is greater than the deterministic length.

The following result shows that an $\alpha$-sufficient value of $q$ leads to the deterministic minimum makespan $make_q$ being a lower bound for the probabilistic minimum makespan $D_\alpha$. Therefore, a lower bound for the deterministic minimum makespan is also a lower bound for the probabilistic minimum makespan.

**Proposition 2** *For a probabilistic JSP, suppose $q$ is $\alpha$-sufficient. Then, for any solution $s$, $\Pr(\mathbf{make}(s) \leq make_q) < 1 - \alpha$. Therefore, $make_q$ is not $\alpha$-achievable, and is a strict lower bound for the $\alpha$-minimum makespan $D_\alpha$, i.e., $D_\alpha > make_q$.*

*Proof:* Since $q$ is $\alpha$-sufficient, there exists a (deterministically) critical path $\pi$ in some (probabilistically) optimal solution $s_o$ with $\Pr(\mathbf{len}(\pi) > len_q(\pi)) > \alpha$. We have $len_q(\pi) = make_q(s_o)$, because $\pi$ is a critical path, and, by definition of $make_q$, we have $make_q(s_o) \geq make_q$. So, $\Pr(\mathbf{len}(\pi) > make_q) > \alpha$. By the definition of makespan, for any sample of the random durations vector, $\mathbf{make}(s_o)$ is at least as a large as $\mathbf{len}(\pi)$. So, we have $\Pr(\mathbf{make}(s_o) > make_q) > \alpha$. Hence, $\Pr(\mathbf{make}(s_o) \leq make_q) = 1 - \Pr(\mathbf{make}(s_o) > make_q) < 1 - \alpha$. This implies $D_\alpha(s_o) > make_q$ since $\Pr(\mathbf{make}(s_o) \leq D_\alpha(s_o)) \geq 1 - \alpha$, by Proposition 1(i). Since $s_o$ is a probabilistically optimal solution, $D_\alpha = D_\alpha(s_o)$, and so $D_\alpha > make_q$. Also, for any solution $s$, we have $D_\alpha(s) \geq D_\alpha > make_q$, so $D_\alpha(s) > make_q$, which implies that $make_q$ is not $\alpha$-achievable using $s$, i.e., $\Pr(\mathbf{make}(s) \leq make_q) < 1 - \alpha$. $\square$

### 4.2.1 Finding $\alpha$-Sufficient $q$-Values

Proposition 2 shows that we can find a lower bound for the probabilistic minimum makespan if we can find an $\alpha$-sufficient value of $q$, and if we can solve (or find a lower bound for) the associated deterministic problem. This section looks at the problem of finding $\alpha$-sufficient values of $q$, by breaking down the condition into simpler conditions.

*In the remainder of Section 4.2, we assume an independent probabilistic JSP.*

Let $\pi$ be some path of some solution. Define $\mu_\pi$ to be $\mathrm{E}[\mathbf{len}(\pi)]$, the expected value of the length of $\pi$ (in the probabilistic JSP), which is equal to $\sum_{A_i \in \pi} \mu_i$. Define $\sigma_\pi^2$ to be $\mathrm{Var}[\mathbf{len}(\pi)]$, the variance of the length of $\pi$, which is equal to $\sum_{A_i \in \pi} \sigma_i^2$, since we are assuming that the durations are independent.

**Defining $\alpha$-adequate $B$.** For $B \geq 0$, write $\theta_B(\pi)$ for $\mu_\pi + B\sigma_\pi$, which equals $\sum_{A_i \in \pi} \mu_i + B\sqrt{\sum_{A_i \in \pi} \sigma_i^2}$. We say that $B$ *is $\alpha$-adequate* if for any (deterministically) critical path $\pi$ of any (probabilistically) optimal solution, $\mathrm{Pr}(\mathbf{len}(\pi) > \theta_B(\pi)) > \alpha$, i.e., there is more than $\alpha$ chance that $\pi$ is more than $B$ standard deviations longer than its expected length.

If each duration is normally distributed, then $\mathbf{len}(\pi)$ will be normally distributed, since it is the sum of independent normal distributions. Even if the durations are not normally distributed, $\mathbf{len}(\pi)$ will often be close to being normally distributed (cf. the central limit theorem and its extensions). So, $\mathrm{Pr}(\mathbf{len}(\pi) > \theta_B(\pi))$ will then be approximately $1 - \Phi(B)$, where $\Phi$ is the unit normal distribution. A $B$ value of slightly less than $\Phi^{-1}(1 - \alpha)$ will be $\alpha$-adequate, given approximate normality.

**Defining $B$-adequate Values of $q$.** We say that $q$ is $B$-adequate if there exists a (deterministically) critical path $\pi$ in some (probabilistically) optimal solution such that $len_q(\pi) \leq \theta_B(\pi)$.

The following proposition shows that the task of finding $\alpha$-sufficient values of $q$ can be broken down. It follows almost immediately from the definitions.

**Proposition 3** *If $q$ is $B$-adequate for some $B$ which is $\alpha$-adequate, then $q$ is $\alpha$-sufficient.*

*Proof:* Since $q$ is $B$-adequate, there exists a (deterministically) critical path $\pi$ in some (probabilistically) optimal solution $s$ such that $len_q(\pi) \leq \theta_B(\pi)$. Since $B$ is $\alpha$-adequate, $\mathrm{Pr}(\mathbf{len}(\pi) > \theta_B(\pi)) > \alpha$, and hence $\mathrm{Pr}(\mathbf{len}(\pi) > len_q(\pi)) > \alpha$, as required. □

**Establishing $B$-adequate Values of $q$.** A value $q$ is $B$-adequate if and only if there exists a (deterministically) critical path $\pi$ in some (probabilistically) optimal solution such that $len_q(\pi) \leq \theta_B(\pi)$, equivalently: $\sum_{A_i \in \pi} \mu_i + q \sum_{A_i \in \pi} \sigma_i \leq \sum_{A_i \in \pi} \mu_i + B\sqrt{\sum_{A_i \in \pi} \sigma_i^2}$, that is, $q \leq B \frac{\sqrt{\sum_{A_i \in \pi} \sigma_i^2}}{\sum_{A_i \in \pi} \sigma_i}$. This can be written as: $q \leq \frac{B}{\sqrt{M_\pi}} \frac{\sqrt{\mathrm{Mean}\{\sigma_i^2 : A_i \in \pi\}}}{\mathrm{Mean}\{\sigma_i : A_i \in \pi\}}$, where $M_\pi$ is the number of activities in path $\pi$, and $\mathrm{Mean}\{\sigma_i : A_i \in \pi\} = \frac{1}{M_\pi}\sum_{A_i \in \pi} \sigma_i$.

If any activity $A_i$ is not uncertain (i.e., its standard deviation $\sigma_i$ equals 0), then it can be omitted from the summations and means. $M_\pi$ then becomes the number of uncertain activities in path $\pi$.

As is well known (and quite easily shown), the root mean square of a collection of numbers is always at least as large as the mean. Hence, $\frac{\sqrt{\text{Mean}\{\sigma_i^2 : A_i \in \pi\}}}{\text{Mean}\{\sigma_i : A_i \in \pi\}}$ is greater than or equal to 1. Therefore, a crude sufficient condition for $q$ to be $B$-adequate is: $q \leq \frac{B}{\sqrt{M}}$, where $M$ is an upper bound for the number of uncertain activities in any path $\pi$ for any probabilistically optimal solution (or we could take $M$ to be an upper bound for the number of uncertain activities in any path $\pi$ for *any* solution). In particular, we could generate $B$-adequate $q$ by choosing $q = \frac{B}{\sqrt{M}}$.

**An $\alpha$-sufficient Value of $q$.** Putting the two conditions together and using Proposition 3, we have that a $q$-value of a little less than $\frac{\Phi^{-1}(1-\alpha)}{\sqrt{M}}$ will be $\alpha$-sufficient, given that the lengths of the paths are approximately normally distributed, where $M$ is an upper bound for the number of uncertain activities in any path $\pi$ for any optimal solution. Hence, by Proposition 2, the minimum makespan $make_q$ of the associated deterministic problem is then a strict lower bound for the $\alpha$-minimum makespan $D_\alpha$. For example, with $\alpha = 0.05$, we have $\Phi^{-1}(1-\alpha) \approx 1.645$ (since there is about 0.05 chance that a normal distribution is more than 1.645 standard deviations above its mean), and so we can set $q$ to be a little less than $\frac{1.645}{\sqrt{M}}$.

One can sometimes generate a larger $\alpha$-sufficient value of $q$, and hence a stronger lower bound $make_q$, by focusing only on the significantly uncertain activities. Choose value $\varepsilon$ between 0 and 1. For any path $\pi$, say that that activity $A_j$ is *$\varepsilon$-uncertain* (with respect to $\pi$) if $\sum \{\sigma_i : A_i \in \pi, \sigma_i \leq \sigma_j\} > \varepsilon \sum \{\sigma_i : A_i \in \pi\}$; then the sum of the durations of the activities which are *not $\varepsilon$-uncertain* is at most a fraction $\varepsilon$ of the sum of all the durations in the path. Hence, the activities in $\pi$ which are not $\varepsilon$-uncertain have relatively small standard deviations. If we define $M_\varepsilon$ to be an upper bound on the number of $\varepsilon$-uncertain activities involved in any path of any (probabilistically) optimal solution, then it can be shown, by a slight modification of the earlier argument, that a $q$-value of $\frac{(1-\varepsilon)B}{\sqrt{M_\varepsilon}}$ will be $B$-adequate, and hence a $q$-value of a little less than $\frac{(1-\varepsilon)\Phi^{-1}(1-\alpha)}{\sqrt{M_\varepsilon}}$ will be $\alpha$-sufficient.

The experiments described in Section 6 use, for varying $n$, problems with $n$ jobs and $n$ activities per job). Solutions which have paths involving very large numbers of activities are unlikely to be good solutions. In particular, one might assume that, for such problems, there will be an optimal solution $s$ and a (deterministically) critical $s$-path $\pi$ involving no more than $2n$ activities. Given this assumption, the following value of $q$ is $\alpha$-sufficient, making $make_q$ a lower bound for the probabilistic minimum makespan: $q = \frac{\Phi^{-1}(1-\alpha)}{\sqrt{2n}}$, e.g., $q = \frac{1.645}{\sqrt{2n}}$ when $\alpha = 0.05$. This motivates the choice of $q_1$ in Table 2 in Section 6.1.

**Probabilistic JSP Example continued.** The number of uncertain activities in our running example (see Section 2.2, Figure 1 and Section 4.1) is 3, so one can set $M = 3$. Using $\alpha = 0.05$, this leads to a choice of $q$ slightly less than $1.645/\sqrt{3} \approx 0.950$. By Proposition 3 and the above discussion, such a value of $q$ is $\alpha$-sufficient. The durations of the associated deterministic problem are given by setting $d_i = \mu_i + q\sigma_i$, and so are $d_1 = 1$, $d_2 = 2 + q/2$, $d_3 = 3 + q/2$, $d_4 = 4 + q/2$ and $d_5 = 5$. Solution $s_a$ is the best solution with makespan $make_q(s_a) = 1 + 5 + (2 + q/2) + (3 + q/2) = 11 + q$. Hence, the minimum deterministic makespan $make_q$ equals approximately 11.95, which is a lower bound for the probabilistic minimum makespan $D_\alpha \approx 12.16$, illustrating Proposition 2.

However, $s_c$ is clearly a poor solution, so we could just consider the other solutions: $\{s_a, s_b, s_d\}$. No (deterministically) critical path of these solutions involves more than two uncertain activities (within the range of interest of $q$-values), so we can then set $M = 2$, and $q = 1.16 \approx 1.645/\sqrt{2}$. This leads to the stronger lower bound of $11 + 1.16 = 12.16$, which is a very tight lower bound for the $\alpha$-minimum makespan $D_\alpha$.

### 4.2.2 Discussion of lower bound

In our example, we were able to use our approach to construct a very tight lower bound for the probabilistic minimum makespan. However, this situation is rather exceptional. Two features of the example which enable this to be a tight lower bound are (a) the best solution has a path which is almost always the longest path; and (b) the standard deviations of the uncertain durations are all equal. In the above analysis, the root mean square is approximated (from below) by the mean. This is a good approximation when the standard deviations are fairly similar, and in an extreme case when the (non-zero) standard deviations of durations are all the same (as in the example), the root mean square is actually equal to the mean.

More generally, there are a number of ways in which our lower bound will tend to be conservative. In particular,

- the choice of $M$ will often have to be conservative for us to be confident that it is a genuine upper bound for the number of uncertain activities in any path for any optimal solution;

- we are approximating a root mean square of standard deviations by the average of the standard deviations: this can be a very crude approximation if the standard deviations of the durations vary considerably between activities;

- we are approximating the random variable **make**($s$) by the random length of a particular path.

The strength of our lower bound method, however, is that it is computationally feasible for reasonably large problems as it uses existing well-developed JSP methods.

## 4.3 Evaluating a Solution Using Monte Carlo Simulation

For a given time value, $D$, we want to assess if there exists a solution for which there is a chance of at most $\alpha$ that its random makespan is greater than $D$. Our methods will all involve generating solutions (or partial solutions), and testing this condition.

As noted earlier, evaluating a solution amounts to solving a PERT problem with uncertain durations, a #P-complete problem (Hagstrom, 1988). As in other #P-complete problems such as the computation of Dempster-Shafer Belief (Wilson, 2000), a natural approach to take is Monte Carlo simulation (Burt & Garman, 1970); we do not try to perform an exact computation but instead choose an accuracy level $\delta$ and require that with a high chance our random estimate is within $\delta$ of the true value. The evaluation algorithm then has optimal complexity (low-degree polynomial) but with a potentially high constant factor corresponding to the number of trials required for the given accuracy.

To evaluate a solution (or partial solution) $s$ using Monte Carlo simulation we perform a (large) number, $N$, of independent trials assigning values to each random variable. Each trial generates a deterministic problem, and we can check very efficiently if the corresponding makespan is greater than $D$; if so, we say that the trial succeeds. The proportion of trials that succeed is then an estimate of $\Pr(\mathbf{make}(s) > D)$, the chance that the random makespan of $s$ is more than $D$. For the case of independent probabilistic JSPs, we can generate the random durations vector by picking, using distribution $P_i$, a value for the random duration $\mathbf{d}_i$ for each activity $A_i$. For the general case, picking a random durations vector will still be efficient in many situations; for example, if the distribution is represented by a Bayesian network.

### 4.3.1 Estimating the Chance that the Random Makespan is Greater than $D$

Perform $N$ trials: $l = 1, \ldots, N$.
For each (trial) $l$:

— Pick a random durations vector using the joint density function.

— Let $T_l = 1$ (the trial *succeeds*) if the corresponding (deterministic) makespan is greater than $D$. Otherwise, set $T_l = 0$.

Let $T = \frac{1}{N} \sum_{l=1}^{N} T_l$ be the proportion of trials that succeed. $T$ is then an estimate of $p$, where $p = \Pr(\mathbf{make}(s) > D)$, the chance that a randomly generated durations vector leads to a makespan (for solution $s$) greater than $D$. The expected value of $T$ is equal to $p$, since $E[T_l] = p$ and so $E[T] = \frac{1}{N} \sum_{l=1}^{N} E[T_l] = p$. The standard deviation of $T$ is $\sqrt{\frac{p(1-p)}{N}}$, which can be shown as follows: $Var[T_l] = E[(T_l)^2] - (E[T_l])^2 = p - p^2 = p(1-p)$. The variables $T_l$ are independent so $Var[T] = \frac{1}{N^2} \sum_{i=1}^{N} Var[T_l] = \frac{p(1-p)}{N} \leq \frac{1}{4N}$. The random variable $NT$ is binomially distributed, and so (because of the deMoivre-Laplace limit theorem (Feller, 1968)) we can use a normal distribution to approximate $T$.

This means that, for large $N$, generating a value of $T$ with the above algorithm will, with high probability, give a value close to $\Pr(\mathbf{make}(s) > D)$. We can choose any accuracy level $\delta > 0$ and confidence level $r$ (e.g., $r = 0.95$), and choose $N$ such that $\Pr(|T - p| < \delta) > r$; in particular, if $r = 0.95$ and using a normal approximation, choosing a number $N$ of trials more than $\frac{1}{\delta^2}$ is sufficient. For fixed accuracy level $\delta$ and confidence level $r$, the number of trials $N$ is a constant: it does not depend on the size of the problem. The algorithm therefore has excellent complexity: the same as the complexity (low-order polynomial) of a single deterministic propagation, and so must be optimal as we clearly cannot hope to beat the complexity of deterministic propagation. However, the constant factor $\frac{1}{\delta^2}$ can be large when we require high accuracy.

### 4.3.2 When is the Solution Good Enough?

Let $D$ be a time value and let $s$ be a solution. Suppose, based on the above Monte-Carlo algorithm using $N$ trials, we want to be confident that $D$ is $\alpha$-achievable using $s$ (i.e., that $\Pr(\mathbf{make}(s) > D) \leq \alpha$). We therefore need the observed $T$ to be at least a little smaller than $\alpha$, since $T$ is (only) an estimate of $\Pr(\mathbf{make}(s) > D)$.

To formalize this, we shall use a confidence interval-style approach. Let $K \geq 0$. Recall that $p = \Pr(\mathbf{make}(s) > D)$ is an unknown quantity that we want to find information about. We say that "$p \geq \alpha$ is $K$-*implausible given the result $T$*" if the following condition holds: $p \geq \alpha$ *implies that $T$ is at least $K$ standard deviations below the expected value*, i.e., $T \leq p - \frac{K}{\sqrt{N}}\sqrt{p(1-p)}$.

If it were the case that $p \geq \alpha$, and "$p \geq \alpha$ is $K$-implausible given $T$", then an unlikely event would have happened. For example, with $K = 2$, (given the normal approximation), such an event will only happen about once every 45 experiments; if $K = 4$ such an event will only happen about once every 32,000 experiments.

If $\Pr(\mathbf{make}(s) > D) \geq \alpha$ is $K$-implausible given the result $T$, then we can be confident that $\Pr(\mathbf{make}(s) > D) < \alpha$: $D$ is $\alpha$-achievable using $s$, so that $D$ is an upper bound of the $\alpha$-minimum makespan $D_\alpha$ and hence of the $\alpha$-minimum makespan $D_\alpha$. The confidence level, based on a normal approximation of the binomial distribution, is $\Phi(K)$, where $\Phi$ is the unit normal distribution. For example, $K = 2$ gives a confidence of around 97.7%.

Similarly, for any $\alpha$ between 0 and 0.5, we say that $p \leq \alpha$ is $K$-implausible given the result $T$ if the following condition holds: $p \leq \alpha$ *implies that $T$ is at least $K$ standard deviations above the expected value, i.e.,* $T \geq p + \frac{K}{\sqrt{N}}\sqrt{p(1-p)}$.

The above definitions of $K$-implausibility are slightly informal. The formal definitions are as follows. Suppose $\alpha \in (0, 0.5]$, $K \geq 0$, $T \in [0, 1]$ and $N \in \{1, 2, \ldots, \}$. We define: $p \geq \alpha$ *is $K$-implausible given $T$* if and only if for all $p$ such that $\alpha \leq p \leq 1$, the following condition holds: $T \leq p - \frac{K}{\sqrt{N}}\sqrt{p(1-p)}$. Similarly, $p \leq \alpha$ *is $K$-implausible given $T$* if and only if for all $p$ such that $0 \leq p \leq \alpha$, the following condition holds: $T \geq p + \frac{K}{\sqrt{N}}\sqrt{p(1-p)}$.

These $K$-implausibility conditions cannot be tested directly using the definition since $p$ is unknown. Fortunately, we have the following result, which gives equivalent conditions that can be easily checked.

**Proposition 4** *With the above definitions:*

   *(i) $p \geq \alpha$ is $K$-implausible given $T$ if and only if $T \leq \alpha - \frac{K}{\sqrt{N}}\sqrt{\alpha(1-\alpha)}$.*

   *(ii) $p \leq \alpha$ is $K$-implausible given $T$ if and only if $T \geq \alpha + \frac{K}{\sqrt{N}}\sqrt{\alpha(1-\alpha)}$.*

*Proof:* (i): If $p \geq \alpha$ is $K$-implausible given $T$, then setting $p$ to $\alpha$ gives $T \leq \alpha - \frac{K}{\sqrt{N}}\sqrt{\alpha(1-\alpha)}$ as required. Conversely, suppose $T \leq \alpha - \frac{K}{\sqrt{N}}\sqrt{\alpha(1-\alpha)}$. The result follows if $K = 0$, so we can assume that $K > 0$. Write $f(x) = (x - T)^2 - \frac{K^2 x(1-x)}{N}$. Now, since $T \leq \alpha - \frac{K}{\sqrt{N}}\sqrt{\alpha(1-\alpha)}$, we have $\alpha > T$ and $(\alpha - T)^2 \geq \frac{K^2\alpha(1-\alpha)}{N}$ so, $f(\alpha) \geq 0$. Also, $f(T) \leq 0$. Since $f(x)$ is a quadratic polynomial with a positive coefficient of $x^2$, this implies that $T$ is either a solution of the equation $f(x) = 0$, or is between the two solutions. Since $f(\alpha) \geq 0$ and $\alpha > T$, it follows that $\alpha$ must either be a solution of $f(x) = 0$, or be greater than the solution(s). This implies, for all $p > \alpha$, $f(p) > 0$, and so $(p - T)^2 > \frac{K^2 p(1-p)}{N}$. Since $p > T$, we have for all $p \geq \alpha$ that $T \leq p - K\sqrt{\frac{p(1-p)}{N}}$, that is, $p \geq \alpha$ is $K$-implausible given $T$, proving (i).

(ii) If $p \leq \alpha$ is $K$-implausible given $T$, then setting $p$ to $\alpha$ gives $T \geq \alpha + K\sqrt{\frac{\alpha(1-\alpha)}{N}}$. Conversely, if $T \geq \alpha + K\sqrt{\frac{\alpha(1-\alpha)}{N}}$, then (since $\alpha \leq 0.5$) $p \leq \alpha$ implies $T \geq p + K\sqrt{\frac{p(1-p)}{N}}$ since the right-hand-side is a strictly increasing function of $p$, so $p \leq \alpha$ is $K$-implausible given $T$, as required. □

Part (i) of this result shows us how to evaluate a solution $s$ with respect to a bound $D$: if we generate $T$ (using a Monte Carlo simulation) which is at least $\frac{K}{\sqrt{N}}\sqrt{\alpha(1-\alpha)}$ less than $\alpha$, then we can have confidence that $p < \alpha$, i.e., $\Pr(\mathbf{make}(s) > D) < \alpha$, and so we can have confidence that $D$ is $\alpha$-achievable using solution $s$, i.e., that $D$ is an upper bound for the probabilistic makespan $D_\alpha(s)$. Part (ii) is used in the branch-and-bound algorithm described in Section 5.2.1, for determining if we can backtrack at a node.

### 4.3.3 GENERATING AN UPPER APPROXIMATION OF THE PROBABILISTIC MAKESPAN OF A SOLUTION

Suppose that, given a solution $s$, we wish to find a time value $D$ which is just large enough such that we can be confident that the probabilistic makespan of $s$ is at most $D$, i.e., that $D$ is an upper bound for the $\alpha$-makespan $D_\alpha(s)$. The Monte Carlo simulation can be adapted for this purpose. We simulate the values of the random makespan $\mathbf{make}(s)$ and record the distribution of these. We decide on a value of $K$, corresponding to the desired degree of confidence (e.g., $K = 2$ corresponds to about 97.7% confidence) and choose $D$ minimal such that the associated $T$ value (generated from the simulation results) satisfies $T \leq \alpha - \frac{K}{\sqrt{N}}\sqrt{\alpha(1-\alpha)}$. Then by Proposition 4(i), $\Pr(\mathbf{make}(s) > D) \geq \alpha$ is $K$-implausible given $T$. We can therefore be confident that $\Pr(\mathbf{make}(s) > D) < \alpha$, so we can have confidence that $D$ is an upper bound for the $\alpha$-makespan $D_\alpha(s)$ of $s$. In the balance of this paper, we will use the notation $D(s)$ to represent our (upper) estimate of $D_\alpha(s)$ found in this way.

## 5. Searching for Solutions

The theoretical framework provides two key tools that we use in building search algorithms. First, we can use Monte Carlo simulation to evaluate a solution or a partial solution (see Section 4.3). Second, with the appropriate choice of a $q$ value, we can solve an associated deterministic problem to find a lower bound on the $\alpha$-minimum makespan for a problem instance (see Section 4.2). In this section, we make use of both these tools (and some variations) to define a number of constructive and local search algorithms. Before describing the algorithms, we recall some of the most important concepts and notation introduced in these earlier sections.

For all of our algorithms, we explicitly deal only with the case of independent probabilistic JSPs where durations are positive integer random variables. Given our approach, however, these algorithms are all valid:

- for the generalized probabilistic case, with the assumptions noted in Section 4, provided we have an efficient way to sample the activity durations;

- for continuous random variables, provided we have a deterministic solver that can handle continuous time values.

## 5.1 Summary of Notation

The remainder of the paper makes use of notation and concepts from earlier sections, which we briefly summarize below.

For a JSP or probabilistic JSP: a solution $s$ totally orders activities requiring the same resource (i.e., activities in the same resource set), so that if activity $A_i$ and $A_j$ require the same resource, then $s$ either determines that $A_i$ must have been completed by the time $A_j$ starts, or vice versa (see Section 2.1). A partial solution partially orders the set of activities in each resource set. Associated with a solution is a non-delay schedule (relative to the solution), where activities without predecessors are started at time 0, and other activities are started as soon as all their predecessors have been completed. The *makespan of a solution* is the time when all jobs have been completed in this associated non-delay schedule. For a probabilistic JSP (see Section 2.2), the makespan $\mathbf{make}(s)$ of a solution $s$ is a random variable, since it depends on the random durations.

The quantity we use to evaluate a solution $s$ is $D_\alpha(s)$, the $\alpha$-makespan of $s$ (also known as the probabilistic makespan of $s$), defined in Section 4.1. The probability that the (random) makespan of $s$ is more than $D_\alpha(s)$ is at most $\alpha$, and approximately equal to $\alpha$. (More precisely, $D_\alpha(s)$ is the smallest time value $D$ such that $\Pr(\mathbf{make}(s) > D)$ is at most $\alpha$.) Value $\alpha$ therefore represents a degree of confidence required. The $\alpha$-minimum makespan $D_\alpha$ (also known as the probabilistic minimum makespan) is the minimum of $D_\alpha(s)$ over all solutions $s$.

A time value $D$ is $\alpha$-achievable using solution $s$ if and only if there is at most $\alpha$ chance that the random makespan is more than $D$. $D$ is $\alpha$-achievable using $s$ if and only if $D \geq D_\alpha(s)$ (see Section 4.1).

Solutions of probabilistic JSPs are evaluated by Monte Carlo simulation (see Section 4.3). A method is derived for generating an "upper approximation" of $D_\alpha$. We use the notation $D(s)$ to represent this upper approximation, which is constructed so that $D(s)$ is approximately equal to $D_\alpha(s)$, and there is a high chance that $D_\alpha(s)$ will be less than $D(s)$—see Section 4.3.3. $D(s)$ thus represents a probable upper bound for the probabilistic minimum makespan.

With a probabilistic job shop problem we often associate a deterministic JSP (see Section 4.2). This mapping is parameterized by a (non-negative real) number $q$. The associated deterministic JSP has the same structure as the probabilistic JSP; the only difference is that the duration of an activity $A_i$ is equal to $\mu_i + q\sigma_i$, where $\mu_i$ and $\sigma_i$ are the mean and standard deviation (respectively) of the duration of $A_i$ in the probabilistic problem. We write $make_q(s)$ for the makespan of a solution $s$ with respect to this associated deterministic JSP, and $make_q$ for the minimum makespan: the minimum of $make_q(s)$ over all solutions $s$. In Section 4.2, it is shown, using Propositions 2 and 3 and the further analysis in Section 4.2.1, that for certain values of $q$, the time value $make_q$ is a lower bound for $D_\alpha$.

## 5.2 Constructive Search Algorithms

Four constructive-search based algorithms are introduced here. Each of them uses constraint-based tree search as a core search technique, incorporating simulation and $q$ values in different ways. In this section, we define each constructive algorithm in detail and then provide a description of the heuristics and constraint propagation building blocks used by each of them.

### 5.2.1 B&B-N: An Approximately Complete Branch-and-Bound Algorithm

Given the ability to estimate the probabilistic makespan of a solution, and the ability to test a condition that implies that a partial solution cannot be extended to a solution with a better probabilistic makespan, an obviously applicable search technique is branch-and-bound (B&B) where we use Monte Carlo simulation to derive both upper- and lower-bounds on the solution quality. If we are able to cover the entire search space, such an approach is approximately complete (only "approximately" because there is always a small probability that we miss an optimal solution due to sampling error).

The B&B tree is a (rooted) binary tree. Associated with each node $e$ in the tree is a partial solution $s_e$, which is a solution if the node is a leaf node. The empty partial solution is associated with the root node. Also associated with each non-leaf node $e$ is a pair of activities, $A_i, A_j, j \neq i$, in the same resource set, whose sequence has not been determined in partial solution $s_e$. The two nodes below $e$ extend $s_e$: one sequences $A_i$ before $A_j$, the other adds the opposite sequence. The heuristic used to choose which sequence to try first is described in Section 5.2.5.

The value of global variable $D^*$ is always such that we have confidence (corresponding to the choice of $K$—see Section 4.3.2) that there exists a solution $s$ whose $\alpha$-makespan, $D_\alpha(s)$, is at most $D^*$. Whenever we reach a leaf node, $e$, we find the upper estimate $D' = D(s_e)$ of the probabilistic makespan $D_\alpha(s)$, by Monte Carlo simulation based on the method of Section 4.3.3. We set $D^* := \min(D^*, D')$. Variable $D^*$ is initialized to some high value.

At non-leaf nodes, $e$, we check to see if it is worth exploring the subtree below $e$. We perform a Monte Carlo simulation for partial solution, $s_e$, using the current value of $D^*$; this generates a result $T$. We use Proposition 4(ii) to determine if $\Pr(\mathbf{make}(s_e) > D^*) \leq \alpha$ is $K$-implausible given $T$; if it is, then we backtrack, since we can be confident that there exists no solution extending the partial solution $s_e$ that improves our current best solution. If $K$ is chosen sufficiently large, we can be confident that we will not miss a good solution.[6]

We refer to this algorithm as *B&B-N* as it performs *B*ranch-and-*B*ound with simulation at each *N*ode.

### 5.2.2 B&B-DQ-L: An Approximately Complete Iterative Tree Search

For an internal node, $e$, of the tree, the previous algorithm used Monte Carlo simulation (but without strong propagation within each trial) to find a lower bound for the probabilistic makespans of all solutions extending partial solution $s_e$. An alternative idea for generating

---

6. Because we are doing a very large number of tests, we need much higher confidence than for a usual confidence interval; fortunately, the confidence associated with $K$ is (based on the normal approximation of a binomial, and the approximation of a tail of a normal distribution) approximately $1 - \frac{1}{K\sqrt{2\pi}} e^{-\frac{1}{2}K^2}$, and so tends to 1 extremely fast as $K$ increases.

**B&B-DQ-L**():
Returns the solution with lowest probabilistic makespan

**1** $(s^*, D^*) \leftarrow$ findFirstB&BSimLeaves$(\infty, 0)$
**2** $q \leftarrow q_{init}$
**3** **while** $q \geq 0$ *AND not timed-out* **do**
**4**     (s, D) $\leftarrow$ findOptB&BSimLeaves$(D^*, q)$
**5**     **if** $s \neq NIL$ **then**
**6**         $s^* \leftarrow s; D^* \leftarrow D$
    **end**
**7**     $q \leftarrow q - q_{dec}$
**end**
**8** return $s^*$

**Algorithm 1**: B&B-DQ-L: An Approximately Complete Iterative Tree Search

such a lower bound is to use the approach of Section 4.2: we find the minimum makespan, over all solutions extending $s_e$, of the associated deterministic problem based on a $q$ value that is $\alpha$-sufficient. This minimum makespan is then (see Proposition 2) a lower bound for the probabilistic makespan. Standard constraint propagation on the deterministic durations enables this lower bound to be computed much faster than the simulation of the previous algorithm. At each leaf node, simulation is used as in B&B-N to find the estimate of the probabilistic makespan of the solution.

This basic idea requires the selection of a $q$ value. However, rather than parameterize this algorithm (as we do with some others below), we choose to perform repeated tree searches with a descending $q$ value.

The algorithm finds an initial solution (line 1 in Algorithm 1) and therefore an initial upper bound, $D^*$, on the probabilistic makespan with $q = 0$. Subsequently, starting with a high $q$ value (one that does not result in a deterministic lower bound), we perform a tree search. When a leaf, $e$, is reached, simulation is used to find $D(s_e)$. With such a high $q$ value, it is likely that the deterministic makespan $make_q(s_e)$ is much greater than $D(s_e)$. Since we enforce the constraint that $make_q(s_e) \leq D(s_e)$, finding $D(s_e)$ through simulation causes the search to return to an interior node, $i$, very high in the tree such that $make_q(S_i) \leq D(s_e)$ where $S_i$ represents the set of solutions in the subtree below node $i$, and $make_q(S_i)$ is the deterministic lower bound on the makespan of those solutions. With high $q$ values, we commonly observed in our experiments that there are only a very few nodes that meet this criterion and, therefore, search is able to very quickly exhaust the search space. When this happens, we reduce the $q$ value by a small amount, $q_{dec}$ (e.g., 0.05), and restart the tree search. Eventually, and often very quickly, we reach a $q$ value such that there exists a full solution, $s_e$, such that $make_q(s_e) \leq D(s_e)$. That solution is stored as the current best and we set $D^* = D(s_e)$. As in B&B-N, $D^*$ is used as an upper bound on all subsequent search.

Algorithm 1 presents pseudocode for the basic algorithm. We make use of two functions not defined using pseudocode:

- findFirstB&BSimLeaves$(c, q)$: creates a JSP with activity durations defined based on the $q$ value passed in and conducts a branch-and-bound search where Monte Carlo simulation is used for each leaf node and standard constraint propagation is used at interior nodes. The first solution that is found whose probabilistic makespan is less than $c$ is returned with the value of its probabilistic makespan. When $c$ is set very high as in line 1, no backtracking is needed to find a solution and therefore only one leaf node is visited and only one simulation is performed.

- findOptB&BSimLeaves($c$, $q$): the same as findFirstB&BSimLeaves($c$, $q$) except the solution with lowest probabilistic makespan is returned rather than the first one found. If no solution is found, a NIL value is returned. Unless the $q$ value is low enough that the deterministic makespan is a lower bound on the probabilistic makespan, this function does not necessarily return the globally optimal solution.

We find a starting solution with $q = 0$ to serve as an initial upper bound on the optimal probabilistic makespan. In practice, B&B-DQ-L is run with a limit on the CPU time. If $q = 0$ is reached within the time limit, this algorithm is approximately complete.

As noted above, it is possible, especially with a high $q$ value, that for a solution, $s_e$, $make_q(s_e)$ is much larger than $D(s_e)$, and therefore the search will backtrack to the deepest interior node such that $make_q(S_i) \leq D(s_e)$. In fact, the assignment of the $D(s_e)$ value is a "global cut" as it is an upper bound on the probabilistic makespan. For technical reasons beyond the scope of this paper, standard constraint-based tree search implementations do not automatically handle such global cuts. We therefore modified the standard behavior to repeatedly post the upper bound constraint on $make_q(S_i)$ causing a series of backtracks up to the correct interior node.

We refer to this algorithm as *B&B-DQ-L* as it does a series of *B*ranch-and-*B*ound searches with *D*escending $q$ values and where simulation is used at the *L*eaves of the tree.

B&B-DQ-L is an example of a novel constraint-based search technique that might be useful in a wider context. When a problem has a cost function that is expensive to evaluate but has an inexpensive, parameterizable lower bound calculation, a search based on over-constraining the problem (i.e., by choosing a parameter value that will not lead to a lower bound) and then iteratively relaxing the bounding function, may be worth investigating. We discuss such an approach in Section 7.

### 5.2.3 B&B-TBS: A Heuristic Tree Search Algorithm

Previous results with an algorithm similar to B&B-N (Beck & Wilson, 2004) indicated that simulation was responsible for a large percentage (e.g., over 95%) of the run-time. We can reduce the number of times we require simulation by only simulating solutions that have a very good deterministic makespan. This *deterministic filtering search* is the central idea for the rest of the algorithms investigated in this paper.

A simple method of filtering solutions is to first spend some fixed amount of CPU time to find a solution, $s_0$, with a low deterministic makespan, $make_q(s_0)$, using a fixed $q$ value and standard constructive tree search. Then, search can be restarted using the same $q$ value and whenever a solution, $s_i$, is found such that $make_q(s_i) \leq make_q(s_0)$, a simulation is run to evaluate $D(s_i)$, our estimate of the probabilistic makespan, $D_\alpha(s_i)$. If the probabilistic makespan found is better than the lowest probabilistic makespan so far, the solution is stored. Search is continued until the entire tree has been explored or the maximum allowed CPU time has expired. Algorithm 2 contains the pseudocode.

**B&B-TBS**($q$):
Returns the solution with lowest probabilistic makespan found

**1** $(s^*, D_{initial}) \leftarrow$ findOptB&B($\infty, q, t_{initial}$)
**2** $D^* \leftarrow \infty$
**3** **while** *solutions exist AND not timed-out* **do**
**4**     $(s, D) \leftarrow$ findNextB&B($D_{initial} + 1, q$, time-remaining)
**5**     $D' \leftarrow$ simulate($s$)
**6**     **if** $D' < D^*$ **then**
**7**         $s^* \leftarrow s; D^* \leftarrow D'$
       **end**
   **end**
**8** return $s^*$

**Algorithm 2**: B&B-TBS: A Heuristic Tree Search Algorithm

As with Algorithm 1, we make use of a number of functions not defined with pseudocode:

- findOptB&B($c$, $q$, $t$): creates a JSP with activity durations defined based on the $q$ value passed in and conducts a deterministic branch-and-bound search for up to $t$ CPU seconds using $c$ as an upper bound on the deterministic makespan. When the search tree is exhausted or the time-limit is reached, the best deterministic solution found (i.e., the one with minimum makespan), together with its deterministic makespan are returned. No Monte Carlo simulation is done.

- findNextB&B($c$, $q$, $t$): this function produces a sequence of solutions (one solution each time it is called) whose deterministic makespan is less than $c$. The problem is defined using the $q$ value and $t$ is the CPU time limit. The solutions produced are the leaves of the B&B search tree in the order encountered by the algorithm. Note that in Algorithm 2, the $c$ value does not change. Given enough CPU time, the algorithm will evaluate the probabilistic makespan of all solutions whose deterministic makespan is less than or equal to $D_{initial}$.

- simulate($s$): our standard Monte Carlo simulation is run on solution $s$ and $D(s)$, the estimate of its probabilistic makespan, $D_\alpha(s)$, is returned.

The algorithm is not complete, even if the choice of $q$ value results in deterministic makespans that are lower bounds on the probabilistic makespan. This is because there is no guarantee that the optimal probabilistic solution will have a deterministic makespan less than $D_{initial}$ and therefore, even with infinite CPU time, it may not be evaluated.

The algorithm is called *B&B-TBS* for *B*ranch-and-*B*ound-*T*imed *B*etter *S*olution: a fixed CPU time is spent to find a good deterministic solution, and then any deterministic solution found that is as good as or better than the initial solution is simulated.

### 5.2.4 B&B-I-BS: An Iterative Heuristic Tree Search Algorithm

A more extreme filtering algorithm first finds an optimal deterministic solution and uses the deterministic makespan as a filter for choosing the solutions to simulate. Using a fixed

**B&B-I-BS**($q$):
Returns the solution with smallest probabilistic makespan found

**1** $(s^*, D_{initial}) \leftarrow \text{findOptB\&B}(\infty, q, t_0 - 1)$
**2** $D^* \leftarrow \text{simulate}(s^*)$
**3** $i \leftarrow 0$
**4** **while** *not timed-out* **do**
**5**     **while** *search is not complete* **do**
**6**         $(s, make_q) \leftarrow \text{findNextB\&B}(D_{initial} \times (1 + i/100) + 1, q, \text{time-remaining})$
**7**         $D \leftarrow \text{simulate}(s)$
**8**         **if** $D < D^*$ **then**
**9**             $s^* \leftarrow s; D^* \leftarrow D$
        **end**
    **end**
**10**     $i \leftarrow i + 1$
  **end**
**11** return $s^*$

**Algorithm 3**: B&B-I-BS: An Iterative Heuristic Tree Search Algorithm

$q$ value, an optimal solution is found and then simulated. If there is CPU time remaining, the search does a series of iterations starting by using the optimal deterministic makespan as the bound. All solutions with a deterministic makespan as good as (or, in general, better than) the current bound are found and simulated. In subsequent iterations, the bound on the deterministic makespan is increased, resulting in a larger set of solutions being simulated. The solution with the lowest estimated probabilistic makespan is returned. On larger problems, an optimal deterministic makespan may not be found within the CPU limit. In such a case, the best deterministic solution that is found is simulated and returned (i.e., only one simulation is done).

More formally, after finding an optimal deterministic solution with makespan, $make_q^*$, a series of iterations beginning with $i = 0$ is executed. For each iteration, the bound on deterministic makespans is set to $make_q^* \times (1+i/100)$. All solutions, $s_e$, whose deterministic makespans, $make_q(s_e) \leq make_q^* \times (1 + i/100)$, are simulated and the one with the lowest probabilistic makespan is returned. Algorithm 3 presents pseudocode which depends on functions defined above.

The algorithm is complete. When $i$ is large enough so that the cost bound is greater than the deterministic makespan of all activity permutations, they will all be simulated. However, $i$ may have to grow unreasonably large and therefore we treat this algorithm as, practically, incomplete.

We refer to this algorithm as *B&B-I-BS* for *B*ranch-and-*B*ound-*I*terative-*B*est *S*olution.

### 5.2.5 HEURISTIC AND CONSTRAINT PROPAGATION DETAILS

The algorithms described above use texture-based heuristics to decide on the pair of activities to sequence and which sequence to try first. The heuristic builds resource profiles that combine probabilistic estimates of the contention that each activity has for each resource and time-point. The maximum point in the resource profiles is selected and an activity pair that contends for the resource at the selected time-point is heuristically chosen. The sequence chosen is the one that maximizes the remaining slack. The intuition is that a pair of activities that is contending for a highly contended-for resource and time-point is a critical pair of activities that should be sequenced early in the search. Otherwise, via constraint propagation from other decisions, the time windows of these activities may be pruned to the point that neither sequence is possible. The texture-based heuristics have a complexity at each search node of $O(mn^2)$ where $m$ is the number of resources and $n$ is the number of activities on each resource.

For a detailed description and analysis of the texture-based heuristic see the work of Beck and Fox (2000) and Beck (1999).

When constraint propagation is used (i.e., all algorithms above except B&B-N), we use the strong constraint propagation techniques in constraint-based scheduling: temporal propagation, timetables (Le Pape, Couronné, Vergamini, & Gosselin, 1994), edge-finder (Nuijten, 1994), and the balance constraint (Laborie, 2003).

## 5.3 Local Search Algorithms

There is no reason why a deterministic filtering search algorithm needs to be based on branch-and-bound. Indeed, given our approach of finding and simulating only solutions with low deterministic makespans, algorithms based on local search may perform better than constructive search algorithms.

In this section, we present two deterministic filtering algorithms based on tabu search.[7] We define each algorithm and then discuss the details of the tabu search procedure itself.

### 5.3.1 TABU-TBS: A TABU SEARCH ANALOG OF B&B-TBS

The central idea behind using tabu search for deterministic filtering search is to generate a sequence of promising deterministic solutions which are then simulated. It seems reasonable to create an analog of B&B-TBS using tabu search. For a fixed $q$ and for a fixed amount $t_{initial}$ of CPU time, at the beginning of a run, a solution with the lowest possible deterministic makespan, $D_{initial}$, is sought. Search is then restarted and whenever a solution, $s$, is found that has a deterministic makespan $make_q(s) \leq D_{initial}$, Monte Carlo simulation is used to approximate the probabilistic makespan. The solution with the lowest estimated probabilistic makespan is returned.

Algorithm 4 presents the pseudocode for this simple approach. We use the following functions (pseudo-code not given):

- findBestTabu$(c, q, t)$: this function is analogous to findOptB&B$(c, q, t)$. Tabu search is run for up to $t$ CPU seconds and the solution with the lowest deterministic makespan (based on the $q$ value) that is less than $c$ is returned.

- findNextTabu$(c, q, t)$: this function is analogous to findNextB&B$(c, q, t)$. A sequence of solutions (one solution each time it is called) whose deterministic makespan is less

---

7. Early experiments explored an even simpler way of using tabu search to solve the probabilistic JSP by incorporating simulation into the neighborhood evaluation. Given a search state, the move operator (see Section 5.3.3 for details) defines the set of neighboring states. For each neighbor, we can run a Monte Carlo simulation and choose the neighbor with the lowest probabilistic makespan. This technique, not surprisingly, proved impractical as considerable CPU time was spent to determine a single move.

**Tabu-TBS**($q$):
Returns the solution with lowest probabilistic makespan found

**1** $(s^*, D_{initial}) \leftarrow \text{findBestTabu}(\infty, q, t_{initial})$
**2** $D^* \leftarrow \infty$
**3** **while** *termination criteria unmet* **do**
**4**     $(s, D) \leftarrow \text{findNextTabu}(D_{initial} + 1, q, \text{time-remaining})$
**5**     $D' \leftarrow \text{simulate}(s)$
**6**     **if** $D' < D^*$ **then**
**7**         $s^* \leftarrow s; D^* \leftarrow D'$
    **end**
  **end**
**8** return $s^*$

**Algorithm 4**: Tabu-TBS: A Local Search Filtering Algorithm

than $c$ is returned. The problem is defined using the $q$ value and $t$ is the CPU time limit. The solution produced is the next solution found by the tabu search that meets the makespan requirement.

We call this algorithm *Tabu-TBS* for *Tabu-T*imed *B*etter *S*olution.

As with B&B-TBS, the $c$ value is not updated in each iteration. The initial search (line 1) is used to find a good deterministic solution and simulation is done on solutions whose deterministic makespan is better than that of the solution found by the initial search.

### 5.3.2 Tabu-I-BS: An Iterative Tabu Search Algorithm

The core tabu search implementation for fixed durations does not necessarily use the entire CPU time (see Section 5.3.3) and, in fact, especially on small instances often terminates very quickly. We can therefore create an iterative tabu-based solver for the probabilistic JSP similar to B&B-I-BS.

In the first phase, using a time limit that is one second less than the overall time limit, tabu search is used to find a very good deterministic solution, based on a fixed $q$ value. That solution is then simulated. Because the tabu search may terminate before the time limit has expired, any remaining time is spent generating solutions with a deterministic makespan within a fixed percentage of the initial solution's deterministic makespan. As with B&B-I-BS, iterations are run with increasing $i$ value starting with $i = 0$. In each iteration, we simulate solutions found by the tabu search whose deterministic makespan is at most $(1 + i/100)D_{initial}$, where $D_{initial}$ is the value of the deterministic makespan found in phase 1. The solution with the lowest probabilistic makespan is returned.[8]

The algorithm is termed *Tabu-I-BS* for *Tabu-I*terative-*B*est *S*earch. The pseudocode for this algorithm is presented in Algorithm 5.

### 5.3.3 Tabu Search Details

The tabu search used to find solutions to problems with deterministic durations is the TSAB algorithm due to Nowicki and Smutnicki (1996). A very restricted move operator (termed

---

8. The *Tabuf* algorithm proposed in Beck and Wilson (2004) corresponds to the first iteration of Tabu-I-BS.

**Tabu-I-BS**($q$):
Returns the solution with smallest probabilistic makespan found

**1** $(s^*, D_{initial}) \leftarrow \text{findBestTabu}(\infty, q, t_0 - 1)$
**2** $D^* \leftarrow \text{simulate}(s^*)$
**3** $i \leftarrow 0$
**4** **while** *not timed-out* **do**
**5**     **while** *termination criteria unmet* **do**
**6**         $(s, make_q) \leftarrow \text{findNextTabu}(D_{initial} \times (1 + i/100) + 1, q, \text{time-remaining})$
**7**         $D \leftarrow \text{simulate}(s)$
**8**         **if** $D < D^*$ **then**
**9**             $s^* \leftarrow s; D^* \leftarrow D$
        **end**
    **end**
**10**     $i \leftarrow i + 1$
  **end**
**11** return $s^*$

**Algorithm 5**: Tabu-I-BS: An Iterative Tabu-based Filtering Algorithm

$N5$ by Blazewicz, Domschke and Pesch, 1996) produces a neighborhood by swapping a subset of the pairs of adjacent activities in the same resource of a given solution. A standard tabu list of ten moves done in the immediate past is kept so as to escape local minima. We use the standard aspiration criteria of accepting move on the tabu list only if the resulting solution is better than *any* solution found so far.

One of the important additions to the basic tabu search mechanism in TSAB is the maintenance of an elite pool of solutions. These are a small set (i.e., 8) of the best solutions that have been encountered so far that is updated whenever a new best solution is encountered. When the standard tabu search stagnates (i.e., it has made a large number of moves without finding a new best solution), search returns to one of the elite solutions and continues search from it. That solution is removed from the set of elite solutions. Search is terminated when either the maximum CPU time is reached or when the elite solution pool is empty.

## 5.4 Summary of Algorithms

Table 1 summarizes the algorithms introduced above.

## 6. Empirical Investigations

Our empirical investigations address two main issues: the scaling behavior of both the approximately complete and heuristic methods as problem size and uncertainty increase and whether using deterministic methods, which represent the uncertainty through duration extensions, is a useful approach. With respect to scaling, there are two interesting subquestions: first, how do the approximately complete techniques compare with each other and, second, is there a cross-over point in terms of problem size above which the heuristic techniques out-perform the approximately complete techniques.

| Name | Deterministic Algorithm | Complete | Description |
|------|------------------------|----------|-------------|
| B&B-N | B&B | Yes | B&B with simulation at each node to find upper and lower bounds |
| B&B-DQ-L | B&B | Yes | B&B with deterministic durations used for lower bounds and simulation is done at each leaf node. The durations decrease in each iteration. |
| B&B-TBS | B&B | No | Find a good deterministic solution, $s$, and restart search, simulating whenever a deterministic solution as good as $s$ is found. |
| B&B-I-BS | B&B | Yes | Find an optimal deterministic solution, $s$. Restart search simulating whenever a deterministic solution within $i\%$ of $s$ is found Repeat with increasing $i$. |
| Tabu-TBS | Tabu | No | Find a good deterministic solution, $s$, and restart search simulating whenever a deterministic solution as good as $s$ is found. |
| Tabu-I-BS | Tabu | No | Find as good a deterministic solution, $s$, as possible. Restart search simulating whenever a deterministic solution within $i\%$ of $s$ is found. Repeat with increasing $i$. |

Table 1: A summary of the algorithms introduced to find the probabilistic makespan for an instance of the job shop scheduling problem with probabilistic durations.

For the heuristic techniques it is necessary to assign fixed durations to each activity. A standard approach is to use the mean duration. However, in such cases there is no representation of the uncertainty surrounding that duration, and this does not take into account that we want a high probability $(1-\alpha)$ of execution. A more general approach is to heuristically use the formulation for the lower bound on $\alpha$-minimum makespans presented in Section 4.2: the duration of activity $A_i$ is defined to be $\mu_i + q\sigma_i$, where $q$ is a fixed non-negative value, and $\mu_i$ and $\sigma_i$ are (respectively) the mean and standard deviation of the duration of $A_i$. Since we are no longer limited to producing a lower bound, we have flexibility in selecting $q$. Intuitively, we want a $q$-value that leads to a situation where good deterministic solutions also have low values of the probabilistic makespan $D_\alpha(s)$. We experiment with a number of $q$-values based on the analysis in Section 4.2 as shown in Table 2. In all cases, we set $B = 1.645$ (see Section 4.2) corresponding to $\alpha = 0.05$. Value $q_3$ was generated for each problem instance by Monte Carlo simulation: simulating 100000 paths of $n$ activities.

## 6.1 Experimental Details

Our empirical investigations examine four sets of probabilistic JSPs of size $\{4 \times 4, 6 \times 6, 10 \times 10, 20 \times 20\}$ (where a $10 \times 10$ problem has 10 jobs each consisting of 10 activities), and for each set, three uncertainty levels $u_j \in \{0.1, 0.5, 1\}$ were considered. A deterministic problem is generated using an existing generator (Watson, Barbulescu, Whitley, & Howe, 2002) with integer durations drawn uniformly from the interval [1, 99]. Three probabilistic instances at different levels of uncertainty are then produced by setting each mean duration $\mu_i$ to be the deterministic duration of activity $A_i$, and by randomly drawing (using a uniform distribution) the standard deviation $\sigma_i$ of the duration of activity $A_i$ from the interval [0, $u_j\mu_i$]. The distribution of each duration is approximately normal. For each problem size, we generate 10 deterministic problems which are transformed into 30 probabilistic instances.

| $q_0$ | $q_1$ | $q_2$ | $q_3$ |
|---|---|---|---|
| 0 | $\frac{1.645}{\sqrt{2n}}$ | $\frac{q_1+q_3}{2}$ | $\frac{1.645}{\sqrt{n}}\frac{\sqrt{\text{Mean}_{A_i\in\pi}\sigma_i^2}}{\text{Mean}_{A_i\in\pi}\sigma_i}$ |

Table 2: The $q$-values used in the experiments. The choices of $q_1$ and $q_3$ are motivated by the analysis in Section 4.2.1.

The problem sizes were chosen to elicit a range of behavior, from the small problems, where the approximately complete algorithms were expected to be able to find and prove (approximate) optimality, to the larger problems, where even the underlying deterministic problems could not be solved to optimality within the time limit used. We chose to use an existing generator rather than, for example, modifying existing benchmark problems, because it allowed us to have full control over the problem structure. The three levels of uncertainty are simply chosen to have low, medium, and high uncertainty conditions under which to compare our algorithms.

Given the stochastic nature of the simulation and the tabu search algorithm, each algorithm is run 10 times on each problem instance with different random seeds. Each run has a time limit of 600 CPU seconds. Each Monte Carlo simulation uses $N = 1000$ independent trials.

The hardware used for the experiments is a 1.8GHz Pentium 4 with 512 MB of main memory running Linux RedHat 9. All algorithms were implemented using ILOG Scheduler 5.3.

Recall that for the B&B-DQ-L algorithm, we employ a descending sequence of $q$ values. For all problems except the 20 × 20 problems, the initial $q$ value, $q_{init}$, was set to 1.25, and the decrement, $q_{dec}$, to 0.05. For the 20 × 20 problems, a $q_{init}$ value of 0.9 was used. This change was made after observing that with $q_{init} = 1.25$, the initial tree search for the 20 × 20 problems would often fail to find a solution or prove that none existed within a reasonable amount of time. We believe this is due to problem instances of that size not having any solution with $q = 1.25$ that satisfied the constraint that the simulated makespan must be less than or equal to the deterministic approximation (i.e., that $make_q(s_e) \leq D(s_e)$—see Section 5.2.2), but yet having a search space that is sufficiently large to require a significant amount of search to prove it. Reducing $q_{init}$ to 0.9 results in an initial solution being found quickly for all instances.

Our primary evaluation criterion is the mean normalized probabilistic makespan ($MNPM$) that each algorithm achieved on the relevant subset of problem instances (we display the data for different subsets to examine algorithm performance for different problem sizes and uncertainty levels). The mean normalized probabilistic makespan is defined as follows:

$$MNPM(a, L) = \frac{1}{|L|} \sum_{l \in L} \frac{D(a,l)}{D_{lb}(l)} \tag{1}$$

where $L$ is a set of problem instances, $D(a, l)$ is our mean estimate of the probabilistic makespan found by algorithm $a$ on $l$ over 10 runs, $D_{lb}(l)$ is the lower bound on the probabilistic makespan for $l$. For all problems except $20 \times 20$, the $D_{lb}$ is found by solving the deterministic problems using $q_1$, a simple, very plausibly $\alpha$-sufficient $q$-value (see Section 4.2 and Table 2). Each instance was solved using constraint-based tree search incorporating the texture-based heuristics and the global constraint propagation used above. A maximum time of 600 CPU seconds was given. All (deterministic) problems smaller than $20 \times 20$ were easily solved to optimality. However, none of the $20 \times 20$ problems were solved to optimality. Because of this, the $D_{lb}$ values were chosen to represent the best solutions found, and so are not true lower bounds.

## 6.2 Results and Analysis

Table 3 presents an overview of the results of our experiments for each problem size and uncertainty level. The results for $q = q_2$ are shown for each heuristic algorithm. There was not a large performance difference among the non-zero $q$-values ($q_1$, $q_2$ and $q_3$). We return to this issue in Section 6.2.2. Each cell in Table 3 is the mean value over 10 independent runs of each of 10 problems. Aside from the $4 \times 4$ instances, all runs reached the 600 CPU second time limit. Therefore, we do not report CPU times.

| Problem Size | Unc. Level | Algorithms | | | | | |
|---|---|---|---|---|---|---|---|
| | | B&B Complete | | B&B Heuristic | | Tabu | |
| | | N | DQ-L | TBS | I-BS | TBS | I-BS |
| $4 \times 4$ | 0.1 | 1.027* | **1.023*** | 1.026 | 1.026 | 1.027 | **1.023** |
| | 0.5 | 1.060* | 1.049* | 1.064 | 1.059 | 1.063 | **1.046** |
| | 1 | 1.151* | 1.129 | 1.154 | 1.149 | 1.153 | **1.128** |
| $6 \times 6$ | 0.1 | 1.034 | **1.021** | 1.022 | 1.022 | 1.027 | 1.023 |
| | 0.5 | 1.113 | **1.073** | 1.083 | 1.077 | 1.074 | 1.074 |
| | 1 | 1.226 | 1.170 | 1.178 | 1.174 | 1.185 | **1.168** |
| $10 \times 10$ | 0.1 | 1.185 | 1.028 | **1.024** | **1.024** | 1.035 | 1.028 |
| | 0.5 | 1.241 | 1.115 | **1.101** | **1.101** | 1.121 | 1.112 |
| | 1 | 1.346 | 1.234 | **1.215** | **1.215** | 1.244 | 1.223 |
| $20 \times 20$† | 0.1 | 1.256 | 1.142 | 1.077 | 1.071 | 1.029 | **1.027** |
| | 0.5 | 1.326 | 1.233 | 1.177 | 1.181 | **1.136** | 1.137 |
| | 1 | 1.482 | 1.388 | 1.334 | 1.338 | **1.297** | 1.307 |

Table 3: The mean normalized probabilistic makespans for each algorithm. '*' indicates a set of runs for which we have, with high confidence, found approximately optimal makespans. '†' indicates problem sets for which normalization was done with approximate lower bounds. The lowest $MNPM$ found for each problem set are shown in bold.

An impression of the results can be gained by looking at the bold entries that indicate the lowest mean normalized probabilistic makespan (MNPM) that was found for each problem set. B&B-N and B&B-DQ-L find approximately optimal solutions only for the smallest problem set, while the B&B-DQ-L and Tabu-I-BS find the lowest probabilistic makespans for both the $4 \times 4$ and $6 \times 6$ problems. Performance of the complete B&B techniques, especially B&B-N, degrade on the $10 \times 10$ problems where the heuristic B&B algorithms find the lowest probabilistic makespans. Finally, on the largest problems, the tabu-based techniques are clearly superior.

One anomaly in the overall results in Table 3 can be seen in the B&B-N and B&B-DQ-L entries for the $4 \times 4$ problems. On two of the three uncertainty levels both algorithms terminate before the limit on CPU time resulting in approximately optimal solutions. However, the mean normalized probabilistic makespans are lower for the B&B-DQ-L algorithm. We conjecture that this is an artifact of the B&B-DQ-L algorithm that biases the simulation toward lower probabilistic makespan values. In B&B-N, a particular solution, $s$, is only simulated once to find $D(s)$. In B&B-DQ-L, the same solution may be simulated multiple times leading to the bias. As an illustration, assume B&B-DQ-L finds an approximately optimal solution $s^*$ while searching the tree corresponding to $q = q' > 0$. On a subsequent iteration with $q = q'' < q'$, provided that the deterministic makespan is less than the previously identified probabilistic makespan (i.e., $make_q(s^*) < D(s^*)$), solution $s^*$ will be found again and simulated again. The actual identity of the current best solution is not used to determine which solutions to simulate. At each subsequent simulation, if a lower value for $D(s^*)$ is generated, it will replace the previous lowest probabilistic makespan value. This leads to a situation where we may re-simulate the same solution multiple times, keeping the lowest probabilistic makespan that is found in any of the simulations. Similar re-simulation is possible with the Tabu-I-BS algorithm.

To test the statistical significance of the results in Table 3, we ran a series of randomized paired-$t$ tests (Cohen, 1995) with $p \leq 0.005$. The results of these statistical tests are displayed in Table 4 for the different problem sizes. The different uncertainty levels have been collapsed so that, for example, the $4 \times 4$ statistics are based on all of the $4 \times 4$ instances. The informal impression discussed above is reflected in these tests with B&B-DQ-L and Tabu-I-BS dominating for the two smallest problem sizes, the branch-and-bound heuristic approaches performing best for the $10 \times 10$ problems, and the tabu-based techniques delivering the best results for the $20 \times 20$ problems.

**Overview.** Our primary interpretation of the performance of the algorithms in these experiments is as follows. For the smaller problems ($4 \times 4$ and $6 \times 6$), the complete techniques are able to cover the entire search space or at least a significant portion of it. Though in the case of B&B-DQ-L, the solutions which are chosen for simulation are heuristically driven by deterministic makespan values, the lower bound results of Section 4.2 ensure that very good solutions will be found provided that iterations with small $q$ values can be run within the CPU time limit. On the $10 \times 10$ problems, the complete techniques are not able to simulate a sufficient variety of solutions as, especially for B&B-N, the heuristic guidance is poor. Note, however, that B&B-DQ-L is competitive with, and, for many problems sets, better than the tabu-based algorithms on the $10 \times 10$ problems. We believe that the $10 \times 10$ results stem from the ability of the B&B heuristic algorithms to quickly find the optimal deterministic solution and then to systematically simulate all solutions with deterministic makespans that are close to optimal. In contrast, the tabu-based algorithms do not systematically enumerate these solutions. Finally, for the largest problems, we hypothesize that tabu search techniques result in the best performance as they are able to find better deterministic solutions to simulate.

| Problem Size | Statistical Significance ($p \leq 0.005$) |
|---|---|
| $4 \times 4$ | {B&B-DQ-L, Tabu-I-BS} < {B&B-TBS, B&B-I-BS, Tabu-TBS, B&B-N} |
| $6 \times 6$ | {B&B-DQ-L, Tabu-I-BS} < {B&B-I-BS} < {B&B-TBS} < {Tabu-TBS} < {B&B-N} |
| $10 \times 10$ | {B&B-TBS, B&B-I-BS} < {Tabu-I-BS, B&B-DQ-L, Tabu-TBS}* < {B&B-N} |
| $20 \times 20$ | {Tabu-TBS, Tabu-I-BS} < {B&B-TBS, B&B-I-BS} < {B&B-DQ-L} < {B&B-N} |

Table 4: The statistically significant relationships among the algorithms for the results shown in Table 3. Algorithms within a set show no significant difference. The '<' relation indicates that the algorithms in the left-hand set have a significantly lower MNPM than the algorithms in the right-hand set. The set indicated by * represents a more complicated relationship amongst the algorithms: Tabu-I-BS < Tabu-TBS but all other pairs in the set show no significant performance differences.

**Problem Size.** As the size of the problems increase, we see the not-unexpected decrease in the quality of the probabilistic makespans found. A simple and reasonable explanation for this trend is that less of the search space can be explored within the given CPU time for the larger problems. There are likely to be other factors that contribute to this trend (e.g., the quality of the lower bound may well systematically decrease as problem size increases).

**Uncertainty Level.** The normalized makespan values also increase within a problem size as the uncertainty level rises. As these results are calculated by normalization against a lower bound, it is possible that the observed decrease in solution quality is actually due to a decrease in the quality of the lower bound rather than a reduction in the quality of the solutions found by the algorithms as uncertainty increases. To test this idea, in Table 5 we normalized the $4 \times 4$ results using the optimal probabilistic makespans found by B&B-N rather than the deterministic lower bound. The table shows that for the algorithms apart from B&B-DQ-L and Tabu-I-BS, the trend of increasing mean normalized probabilistic makespan is still evident. For these algorithms, at least, the putative decreasing quality of the lower bound cannot be the entire explanation for the trend of worse performance results for higher levels of uncertainty. In Section 6.2.2, we revisit this question and provide evidence that could explain why the algorithms perform worse when uncertainty is increased.

These results also lend credibility to the conjecture that the observed "super-optimal" performance of B&B-DQ-L and Tabu-I-BS on the small problems is due to repeatedly simulating the same solution. At low levels of uncertainty, repeated simulations of the truly best solution will not vary greatly, resulting in a MNPM value of about 1. With higher levels of uncertainty, the distribution of simulated makespans is wider and, therefore, repeated simulation of the same solution biases results toward smaller probabilistic makespan values. This is what we observe in the results of B&B-DQ-L and Tabu-I-BS in Table 5.

| Unc. | Algorithms | | | | | |
|------|------------|---|---|---|---|---|
| | B&B Complete | | B&B Heuristic | | Tabu | |
| Level | N | DQ-L | TBS | I-BS | TBS | I-BS |
| 0.1 | 1.004 | 1.000 | 1.003 | 1.002 | 1.003 | 0.999 |
| 0.5 | 1.008 | 0.998 | 1.012 | 1.008 | 1.011 | 0.995 |
| 1 | 1.015 | 0.996 | 1.018 | 1.013 | 1.017 | 0.996 |

Table 5: The mean normalized probabilistic makespans for each algorithm on the $4 \times 4$ problem set normalized by the optimal probabilistic makespans found by B&B-N.

In the balance of this section, we turn to more detailed analysis of the algorithms.

### 6.2.1 Analysis: B&B Complete Algorithms

The performance of B&B-N is poor when it is unable to exhaustively search the branch-and-bound tree. The high computational cost of running simulation at every node and the relatively weak lower bound that partial solutions provide[9] conspire to result in a technique that does not scale beyond very small problems.

| Problem | Uncertainty Level | | |
|---------|------|------|------|
| Size | 0.1 | 0.5 | 1 |
| $4 \times 4$ | 0 | 0 | 0 |
| $6 \times 6$ | 0 | 0.5 | 0.75 |
| $10 \times 10$ | 0.95 | 0.85 | 0.9 |
| $20 \times 20$ | 0.9 | 0.9 | 0.9 |

Table 6: The lowest $q$ value used for each problem size and uncertainty level for the B&B-DQ-L. For all problems except $20 \times 20$, the initial $q$ value is 1.25. For the $20 \times 20$ problems, the initial $q$ value is 0.9

B&B-DQ-L is able to perform somewhat better than B&B-N on larger problems even when it is not able to exhaustively search each tree down to $q = 0$. Table 6 shows the minimum $q$ values attained for each problem size and uncertainty level. The deterministic durations defined by the $q$ value serve to guide and prune the search for each iteration and, therefore, as with the heuristic algorithms (see below), the search is heuristically guided to the extent that solutions with low deterministic makespans also have low probabilistic makespans. However, the characteristics of the solutions found by the search are unclear. Recall that B&B-DQ-L starts with a high $q$ value that, in combination with the constraint that the deterministic makespan must be less than or equal to the best simulated probabilis-tic makespan found so far, significantly prunes the search space. Ideally, we would like the search with high $q$ to find solutions with very good probabilistic makespans both because we wish to find good solutions quickly and because the simulated probabilistic makespan values are used to prune subsequent search with lower $q$ values. Therefore, in an effort to better understand the B&B-DQ-L search, we examine the characteristics of the initial solutions it finds.

---

9. One idea for improving the lower bound that we did not investigate is to incorporate the resource-based propagators (e.g., edge-finding) into the evaluation of a partial solution. In a single trial at an internal node, the deterministic makespan is found by sampling from the distributions and then finding the longest path in the temporal network. After the sampling, however, it is possible to apply the standard propagation techniques which might insert additional edges into the precedence graph and thereby increase the makespan, improving the lower bound.

Some idea of the quality of the solutions produced by high $q$ values can be seen by comparing the probabilistic makespan found with high $q$ (the first solution found) with the best solution found in that run. Table 7 presents this comparison in the form of $D_f$, the mean normalized makespans for the initial solutions found by B&B-N and B&B-DQ-L. These data indicate that the first solution found by B&B-DQ-L is much better than that found by B&B-N. When B&B-N searches for its initial solution, the upper bound on the deterministic makespan does not constrain the problem: a solution is therefore very easy to find (i.e., with no backtracking) but there is little constraint propagation or heuristic information available to guide the search to a solution with a small makespan. In contrast, when B&B-DQ-L searches for an initial solution, the high $q$ value means that it is searching in a highly constrained search space because the deterministic makespan must be less than the probabilistic makespan. Therefore, there is a very tight upper bound on the deterministic makespan (relative to the durations that incorporate the $q$ values). In many cases, the initial iterations fail to find any feasible solutions, but do so very quickly. Eventually, the $q$ value is low enough to allow a feasible solution, however the search for that solution is strongly guided by propagation from the problem constraints. In summary, the initial search for B&B-N has no guidance from the constraint propagation toward a good solution while that of B&-DQ-L is guided by constraint propagation in an overly constrained problem. Table 7 shows that, in these experiments, such guidance tends to result in better initial solutions. We believe that this observation may be useful more generally in constraint solving (see Section 7).

To provide a fuller indication of the performance differences, Table 7 also presents the improvement over the first solution that is achieved: the difference between the first solution and the last solution ($D_l$) found by each algorithm ($D_l$ is the value reported in Table 3). On the larger problem sets, the improvement made on the first solution by B&B-DQ-L is greater. For the smaller problem sets, the improvement by B&B-N is greater than for B&B-DQ-L, however, we suspect a ceiling effect reduces the amount that B&B-DQ-L can improve (i.e., the initial solutions are already quite close to optimal).

### 6.2.2 Analysis: Heuristic Algorithms

We now turn to the performance of the heuristic algorithms. We first examine the hypothesis that their performance is dependent on two factors: the ability of the algorithms to find solutions with low deterministic makespans and the correlation between good deterministic and probabilistic makespans. Then we turn to an analysis of the effect of the differing $q$ values on the heuristic algorithm performance.

**Finding Good Deterministic Makespans.** It was argued above that the performance of the heuristic techniques (and B&B-DQ-L) is dependent upon the ability to find solutions with good deterministic makespans. To provide evidence for this argument, we looked at the quality of the best deterministic solutions found by B&B-I-BS and Tabu-I-BS. We hypothesize that the better performing algorithm will also have found better deterministic solutions than the worse performer.

| Problem Size | Unc. Level | B&B-N | | B&B-DQ-L | |
|---|---|---|---|---|---|
| | | $D_f$ | $D_f - D_l$ | $D_f$ | $D_f - D_l$ |
| 4 × 4 | 0.1 | 1.089 | 0.062 | 1.028 | 0.005 |
| | 0.5 | 1.119 | 0.059 | 1.078 | 0.029 |
| | 1 | 1.227 | 0.076 | 1.165 | 0.036 |
| 6 × 6 | 0.1 | 1.106 | 0.072 | 1.067 | 0.046 |
| | 0.5 | 1.163 | 0.050 | 1.108 | 0.035 |
| | 1 | 1.301 | 0.075 | 1.221 | 0.051 |
| 10 × 10 | 0.1 | 1.191 | 0.006 | 1.069 | 0.045 |
| | 0.5 | 1.258 | 0.017 | 1.151 | 0.050 |
| | 1 | 1.369 | 0.005 | 1.269 | 0.054 |
| 20 × 20 | 0.1 | 1.259 | 0.003 | 1.168 | 0.026 |
| | 0.5 | 1.332 | 0.004 | 1.242 | 0.009 |
| | 1 | 1.494 | 0.008 | 1.404 | 0.016 |

Table 7: The mean normalized makespan for the first solutions found by each algorithm ($D_f$) and the difference between the mean normalized makespans of the first and last solutions ($D_f - D_l$).

Table 8 presents results for each algorithm on the two largest problem sets.[10] The mean normalized deterministic makespan ($MNDM$) is calculated as follows:

$$MNDM(a, L) = \frac{1}{|L|} \sum_{l \in L} \frac{make_q(a, l)}{make_{q,min}(l, B\&B - I - BS)} \qquad (2)$$

where $L$ is a set of problem instances, $make_q(a, l)$ is the mean deterministic makespan found by algorithm $a$ on $l$ over 10 runs, $make_{q,min}(l, B\&B - I - BS)$ is the lowest deterministic makespan found by the B&B-I-BS algorithm over all runs on problem $l$. MNDM, therefore, provides a relative measure of the quality of the average deterministic makespans from the two algorithms: the higher the value, the worse the average makespan found relative to B&B-I-BS.

Table 8 is consistent with our hypothesis. On the 10 × 10 problems, where B&B-I-BS outperforms Tabu-I-BS, the former is able to find solutions with a lower mean deterministic makespan. For the 20 × 20 problems the results are reversed with Tabu-I-BS finding both better mean deterministic makespans and better probabilistic makespans.

This result lends support to the original motivation for the deterministic filtering algorithms: the performance of these algorithms in terms of probabilistic solution quality is positively related to the quality of the deterministic solutions they are able to find. The next section addresses the question of why this performance relationship is observed.

---

10. We show only the 10 × 10 and 20 × 20 problems sets as they are not influenced by the conjectured repeated simulation behavior of Tabu-I-BS.

| Problem | Uncertainty | MNDM | |
|---|---|---|---|
| Size | Level | B&B-I-BS | Tabu-I-BS |
| | 0.1 | 1.000 | 1.002 |
| $10 \times 10$ | 0.5 | 1.000 | 1.004 |
| | 1 | 1.000 | 1.004 |
| | 0.1 | 1.045 | 1.002 |
| $20 \times 20$ | 0.5 | 1.041 | 0.998 |
| | 1 | 1.037 | 1.002 |

Table 8: The mean normalized deterministic makespan (MNDM) for B&B-I-BS and Tabu-I-BS.

**The Correlation Between Deterministic and Probabilistic Makespan.** The ability of the algorithms to find good deterministic makespans would be irrelevant to their ability to find good probabilistic makespans without some correlation between the two. It is reasonable to expect that the level of uncertainty in a problem instance has an impact on this correlation: at low uncertainty the variations in duration are small, meaning that we can expect the probabilistic makespan to be relatively close to the deterministic makespan. When the uncertainty level is high, the distribution of probabilistic makespans for a single solution will be wider, resulting in less of a correlation. We hypothesize that this impact of uncertainty level contributes to the observed performance degradation (see Tables 3 and 5) of the heuristic techniques with higher uncertainty levels as problem size is held constant.

To examine our hypothesis we generated 100 new $10 \times 10$ deterministic JSP problem instances with the same generator and parameters used above. The standard deviations for the duration of each activity in the 100 instances were generated independently for each of five uncertainty levels $u_j \in \{0.1, 0.5, 1, 2, 3\}$ resulting in a total of 500 problem instances (100 for each uncertainty level). For each instance and for each of the four $q$ values (as in Table 2), we then randomly generated 100 deterministic solutions which were then simulated. Using the R statistical package (R Development Core Team, 2004), we measured the correlation coefficient for each problem set. Each cell in Table 9 is the result of 10000 pairs of data points: the deterministic and probabilistic makespans for 100 random deterministic solutions for each of 100 problem instances.

| Uncertainty Level | $q_0$ | $q_1$ | $q_2$ | $q_3$ |
|---|---|---|---|---|
| 0.1 | 0.9990 | 0.9996 | 0.9996 | 0.9995 |
| 0.5 | 0.9767 | 0.9912 | 0.9917 | 0.9909 |
| 1 | 0.9176 | 0.9740 | 0.9751 | 0.9736 |
| 2 | 0.8240 | 0.9451 | 0.9507 | 0.9517 |
| 3 | 0.7381 | 0.9362 | 0.9418 | 0.9423 |

Table 9: The correlation coefficient ($r$) comparing pairs of deterministic and probabilistic makespans for a set of $10 \times 10$ probabilistic JSPs. Each cell represents the correlation coefficient for 10000 deterministic, probabilistic pairs.

Table 9 supports our explanation for the performance of the heuristic techniques. As the uncertainty level increases, the correlation between the deterministic makespan and corresponding probabilistic makespan lessens. The strength of the correlation is somewhat surprising: even for the highest uncertainty level where the standard deviation of the duration of an activity is uniformly drawn from between 0 and 3 times its mean duration, the correlation is above 0.94 for $q_2$ and $q_3$. This is a positive indication for the heuristic algorithms as it suggests that they may scale well to higher uncertainty levels provided a reasonable $q$ value is used. We examine the impact of the $q$ values in the original experiments and the implications of the deterministic/probabilistic makespan correlation in the next section.

It should be emphasized that these results are based on correlations between deterministic and probabilistic makespans for randomly generated solutions. We have not addressed how these correlations might change for high-quality solutions, which might be considered as a more appropriate population from which to sample. One technical difficulty for the design of an experiment to examine this, is to ensure a sufficiently randomized sample from the population of "good" solutions; also, the result could depend strongly on the (rather arbitrary) particular choice of quality cutoff for solutions.

**The Effect of the $q$ Values.** Each of the heuristic algorithms requires a fixed $q$ value.[11] We experimented with four different values (see Table 2). Table 10 displays the significant pairwise differences among the $q$ values for each heuristic as measured by randomized paired-$t$ tests (Cohen, 1995) with $p \leq 0.005$. As can be observed, there are almost no significant differences for low levels of uncertainty (0.1 or 0.5) or for the smallest problem set. For higher levels of uncertainty and larger problems, using $q_0$ is never better than using one of the higher $q$ values and in many cases, $q_0$ results in the worst mean makespan. Among the other $q$-values, for the majority of the problem sets and algorithms there are no significant differences. For a given algorithm, it is never the case that a lower $q$ value leads to significantly better results than a higher $q$ value.

The correlation results in Table 9 provide an explanation for these differences. For the $10 \times 10$ problems, the performance of the $q_0$ algorithms is competitive when there is not a large difference in the correlations between deterministic and probabilistic solutions (i.e., at uncertainty levels 0.1 and 0.5). When the uncertainty level is 1, there is a significant reduction in the correlation coefficient for $q_0$ and a corresponding reduction in the mean normalized probabilistic makespans found by the algorithms using $q_0$.

### 6.3 Summary

The results of our experiments can be summarized as follows:

- The most principled use of simulation (B&B-N) is only useful on small problems. The simulation time is a major component of the run-time resulting in very little exploration of the search space.

- Algorithm B&B-DQ-L, based on the idea of iteratively reducing a parameter that determines the validity of the lower bound, results in equal performance on small prob-lems and much better performance on larger problems when compared to B&B-N. More work is needed to understand the behavior of the algorithm, however preliminary evidence indicates that it is able to find good solutions quickly in the current application domain.

---

11. We are not addressing the behavior of B&B-DQ-L, where the $q$ descends *during* the run of the algorithm. We are only examining the algorithms with fixed $q$ values.

| Problem Size | Unc. Level | B&B | | Tabu | |
|---|---|---|---|---|---|
| | | TBS | I-BS | TBS | I-BS |
| $4 \times 4$ | 0.1 | - | - | - | $q_2 < q_1$ |
| | 0.5 | - | - | - | - |
| | 1 | - | - | - | - |
| | ALL | - | - | - | - |
| $6 \times 6$ | 0.1 | - | - | - | - |
| | 0.5 | - | - | - | - |
| | 1 | $q_2 < \{q_1, q_3\} < q_0$ | - | $\{q_1, q_2, q_3\} < q_0$ | - |
| | ALL | $q_2 < \{q_0, q_1\}$ | - | $\{q_1, q_2, q_3\} < q_0$ | $q_1 < q_0$ |
| $10 \times 10$ | 0.1 | - | - | - | - |
| | 0.5 | - | - | - | - |
| | 1 | $\{q_1, q_2, q_3\} < q_0$ | $\{q_1, q_2, q_3\} < q_0$ | $\{q_1, q_3\} < q_0$ | $\{q_2, q_3\} < q_0$ |
| | ALL | $\{q_1, q_2, q_3\} < q_0$ | $\{q_1, q_2, q_3\} < q_0$ | $q_1 < q_0$ | $\{q_1, q_2, q_3\} < q_0$ |
| $20 \times 20$ | 0.1 | - | - | - | $q_2 < q_0$ |
| | 0.5 | - | - | - | - |
| | 1 | $\{q_1, q_2\} < q_0$ | $\{q_2, q_3\} < q_0$ $q_2 < q_1$ | $\{q_1, q_2, q_3\} < q_0$ | $\{q_1, q_2, q_3\} < q_0$ |
| | ALL | $q_2 < q_1 < q_0$ $q_3 < q_0$ | $\{q_2, q_3\} < q_0$ | $\{q_1, q_2, q_3\} < q_0$ | $\{q_1, q_2, q_3\} < q_0$ |

Table 10: The results of pair-wise statistical tests for each algorithm and problem set. The notation $a < b$ indicates that the algorithm using $q = a$ achieved a significantly better solution (i.e., lower probabilistic makespan) than when it used $q = b$. '-' indicates no significant differences. All statistical tests are randomized paired-$t$ tests (Cohen, 1995) with $p \leq 0.005$.

- A series of heuristic algorithms were proposed based on using deterministic makespan to filter the solutions which would be simulated. It was demonstrated that the performance of these algorithms depends on their ability to find good deterministic makespans and the correlation between the quality of deterministic and probabilistic solutions. It was shown that even for problems with a quite high uncertainty level, deterministic problems can be constructed that lead to a strong deterministic/probabilistic makespan correlation.

- Central to the success of the heuristic algorithms was the use of a $q$ value that governed the extent to which duration uncertainty was represented in the durations of activities in deterministic problems. It was shown that such an incorporation of uncertainty data leads to a stronger correlation between deterministic and probabilistic makespans and the corresponding ability to find better probabilistic makespans.

# 7. Extensions and Future Work

In this section, we look at three kinds of extensions of this work. First, we show that our theoretical framework in fact applies to far more general probabilistic scheduling problems than just job shop scheduling. In Section 7.2, we discuss ways in which the algorithms for probabilistic JSP presented in this paper might be improved. Finally, we discuss the possibility of developing the central idea in the B&B-DQ-L algorithm into a solving approach for general constraint optimization problems.

## 7.1 Generalization to Other Scheduling Problems

The results in this paper have been derived for the important case of job shop scheduling problems. In fact, they are valid for a much broader class of scheduling problems, including resource-constrained project scheduling problems of a common form (e.g., a probabilistic version of the deterministic problems studied in the work of Laborie, 2005). In this section, we describe how to extend our framework and approaches.

Our approach relies on the fact that in the job shop scheduling problem, one can focus on orderings of activities, rather than directly on assignments of start times for activities; specifically, the definition of minimum makespan based on orderings is equivalent to the one based on start time assignments; this equivalence holds much more generally.

First, in 7.1.1, we give some basic definitions and properties which are immediate extensions of those defined in Section 2. Then, in 7.1.2, we characterize a class of scheduling problems which have the properties we require, by use of a logical expression to represent the constraints of the problem. In 7.1.3 we give the key result relating the schedule-based minimum makespan with the ordering-based minimum makespan. Section 7.1.4 discusses the extended class of probabilistic scheduling problems, and Section 7.1.5 considers different optimization functions.

### 7.1.1 SCHEDULES, ORDERINGS AND MAKESPANS

As in Section 2, we are given a set $\mathcal{A}$ of activities, where activity $A_i \in \mathcal{A}$ has an associated positive duration $d_i$ (for the deterministic case). A *schedule* (for $\mathcal{A}$) is defined to be a function from the set of activities to the set of time-points (which are non-negative numbers), defining when each activity starts. Let $Z$ be a schedule. The makespan $make(Z)$ of schedule $Z$ is defined to be the time at which the last activity has been completed, i.e., $\max_{A_i \in \mathcal{A}}(Z(A_i) + d_i)$. We say that $Z$ *orders* $A_i$ *before* $A_j$ if and only if $A_j$ starts no earlier than $A_i$ ends, i.e., $Z(A_i) + d_i \leq Z(A_j)$.

An essential aspect of job shop problems and our approach is that one can focus on orderings of the activities rather than on schedules; in Section 2 we use the term *solution* for an ordering that satisfies the constraints of a given JSP. Define an *ordering* (on $\mathcal{A}$) to be a strict partial order on $\mathcal{A}$, i.e., an irreflexive and transitive relation on the set of activities. Hence, for ordering $s$, for all $A_i \in \mathcal{A}$, $(A_i, A_i) \notin s$, and if $(A_i, A_j) \in s$ and $(A_j, A_k) \in s$, then $(A_i, A_k) \in s$. If $(A_i, A_j) \in s$, then we say that $s$ *orders* $A_i$ *before* $A_j$; we also say that $A_i$ is a *predecessor* of $A_j$. A *path* in $s$ (or an *s-path*) is a sequence of activities such that if $A_i$ precedes $A_j$ in the sequence, then $s$ orders $A_i$ before $A_j$. The length $len(\pi)$ of a path $\pi$ (in an ordering) is defined to be the sum of the durations of the activities in the path, i.e., $\sum_{A_i \in \pi} d_i$. The *makespan*, *make(s)*, of an ordering $s$ is defined to be the length of a longest $s$-path. An $s$-path $\pi$ is said to be a *critical s-path* if the length of $\pi$ is equal to the makespan of the ordering $s$, i.e., it is one of the longest $s$-paths.

Any schedule has an associated ordering. For schedule $Z$ define the ordering *sol(Z)* as follows: *sol(Z)* orders activity $A_i$ before $A_j$ if and only if $Z$ orders $A_i$ before $A_j$.

Conversely, from an ordering one can define a non-delay schedule, which is optimal among schedules compatible with the ordering, by starting each activity as soon as its predecessors finish. Let $s$ be an ordering. We inductively define schedule $Z = sched(s)$ as follows: if $A_i$ has no predecessor, then we start $A_i$ at time 0, i.e., $Z(A_i) = 0$. Otherwise, we set $Z(A_i) = \max_{A_j \in pred(A_i)}(Z(A_j) + d_j)$, where $pred(A_i)$ is the set of predecessors of $A_i$. The fact that $s$ is acyclic guarantees that this defines a schedule. As in Section 2.1, we have the following two important properties. The first states that the makespan of an ordering is equal to the makespan of its associated schedule. The second states that the makespan of a schedule is no better than the makespan of its associated ordering.

**Proposition 5**

*(i) For any ordering $s$, $make(sched(s)) = make(s)$.*

*(ii) For any schedule $Z$, $make(sol(Z)) \leq make(Z)$.*

The proof of these is straight-forward. It follows easily by induction that if a schedule $Z$ respects the precedence constraints expressed by an ordering $s$, then the last activity of any $s$-path can end no earlier in $Z$ than the length of the path; applying this to a critical path implies (ii) $make(sol(Z)) \leq make(Z)$, and implies half of (i): $make(sched(s)) \geq make(s)$. By working backwards from an activity that finishes last in $sched(s)$, and choosing an immediate predecessor at each stage, one generates (in reverse order) a path in $s$ whose length is equal to $make(sched(s))$, hence showing that $make(sched(s)) \leq make(s)$, and proving (i).

### 7.1.2 POSITIVE PRECEDENCE EXPRESSIONS

We will define a class of scheduling problems, using what we call positive precedence expressions (PPEs) to represent the constraints. Each of these scheduling problems assumes no preemption (so activities cannot be interrupted once started) and we will again use makespan as the cost function.

For activities $A_i$ and $A_j$, the expression *before(i, j)* is interpreted as the constraint (on possible schedules) that activity $A_j$ starts no earlier than the end of activity $A_i$. Such expressions are called *primitive precedence expressions*. A *positive precedence expression* is defined to be a logical formula built from primitive precedence expressions, conjunctions and disjunctions. (The term "positive" is used since they do not involve negations.) Formally, the set $\mathcal{E}$ of positive precedence expressions (over $\mathcal{A}$) is defined to be the smallest set such that (a) $\mathcal{E}$ contains *before(i, j)* for each $A_i$ and $A_j$ in $\mathcal{A}$, and (b) if $\varphi$ and $\psi$ are in $\mathcal{E}$, then $(\varphi \wedge \psi)$ and $(\varphi \vee \psi)$ are both in $\mathcal{E}$.

Positive precedence expressions over $\mathcal{A}$ are interpreted as constraining schedules on $\mathcal{A}$. Let $\varphi \in \mathcal{E}$ be a PPE and let $Z$ be a schedule. We define "$Z$ satisfies $\varphi$" recursively as follows:

- $Z$ satisfies primitive precedence expression $before(i, j)$ if and only if $Z$ orders $A_i$ before $A_j$, i.e., if $Z(A_i) + d_i \leq Z(A_j)$;

- $Z$ satisfies the conjunction of two constraint expressions if and only if it satisfies both of them;

- $Z$ satisfies the disjunction of two constraint expressions if and only if it satisfies at least one of them.

Similarly, for ordering $s$ and positive precedence expression $\varphi$ we can recursively define "$s$ satisfies $\varphi$" in the obvious way: $s$ satisfies $before(i, j)$ if and only if $s$ orders $A_i$ before $A_j$. Ordering $s$ satisfies $(\varphi \wedge \psi)$ if and only if it satisfies both $\varphi$ and $\psi$. Ordering $s$ satisfies $(\varphi \vee \psi)$ if and only if it satisfies either $\varphi$ or $\psi$.

Positive precedence expressions are powerful enough to represent the constraints of a job shop scheduling problem, or of a resource-constrained project scheduling problem.

**JSPs as Positive Precedence Expressions.** Resource constraints in a job shop scheduling problem give rise to disjunctions of primitive precedence expressions: for each pair of activities $A_i$ and $A_j$ which require the same resource, the expression $before(i, j) \vee before(j, i)$ which expresses that $A_i$ and $A_j$ do not overlap (one of them precedes the other). The ordering of activities in a job can be expressed in terms of primitive expressions: $before(i, j)$ when $A_i$ precedes $A_j$ within some job. Hence, the constraints in a job shop problem can be expressed as a positive precedence expression in conjunctive normal form, i.e., a conjunction of disjunctions of primitive precedence expressions.

**RCPSPs as PPEs.** The constraints in a resource-constrained project scheduling problem (RCPSP) (Pinedo, 2003; Brucker et al., 1999; Laborie & Ghallab, 1995; Laborie, 2005) can also be expressed as a positive precedence expression in conjunctive normal form. In an RCPSP, we have precedence constraints between activities, each of which can be expressed as a primitive precedence expression; let $\varphi$ be the conjunction of these. In a RCPSP, there are again a set of resources, each with a positive capacity. Associated with each activity $A_i$ and resource $r$ is the rate of usage $A_i(r)$ of resource $r$ by activity $A_i$. We have the following resource constraints on a schedule: for each resource $r$, at any time-point $t$, the sum of $A_i(r)$ over all activities $A_i$ which are in progress at $t$ (i.e., which have started by $t$ but not yet ended) must not exceed the capacity of resource $r$.

Define a forbidden set (or conflict set) to be a set of activities whose total usage of some resource exceeds the capacity of the resource. Let $\mathcal{F}$ be the set of forbidden sets. (If we wished, we could delete from $\mathcal{F}$ any set which is a superset of any other set in $\mathcal{F}$; we could also delete any set $H$ which contains elements $A_i$ and $A_j$ such that $A_i$ precedes $A_j$ according to $\varphi$.) The resource constraints can be expressed equivalently as: for all $H \in \mathcal{F}$, there exists no time at which every activity in $H$ is in progress. This holds if and only if for each $H \in \mathcal{F}$, there exist two activities in $H$ which do not overlap (since if all pairs of activities in $H$ overlap then all activities in $H$ are in progress at the latest start time of activities in $H$), i.e., there exists $A_i, A_j \in H$ with $before(i, j)$. Hence, a schedule satisfies the resource constraints if and only if it satisfies the positive precedence expression $\psi$ defined

to be

$$\bigwedge_{H \in \mathcal{F}} \bigvee_{\substack{A_i, A_j \in H \\ i \neq j}} \textit{before}(i, j).$$

Therefore, expression $(\varphi \wedge \psi)$ represents the RCPSP, i.e., a schedule satisfies the constraints of the RCPSP if and only if it satisfies $(\varphi \wedge \psi)$.

Another class of scheduling problems, each of which can be represented by a positive precedence expression, is the class based on AND/OR precedence constraints (Gillies & Liu, 1995; Möhring, Skutella, & Stork, 2004).

### 7.1.3 Solutions and Minimum Makespan

For a fixed positive precedence expression $\varphi$ over $\mathcal{A}$, we say that schedule $Z$ is *valid* if it satisfies $\varphi$. We say that ordering $s$ is a *solution* if it satisfies $\varphi$. If ordering $s$ satisfies $\textit{before}(i, j)$, then, by construction, $\textit{sched}(s)$ satisfies $\textit{before}(i, j)$. It also follows immediately that schedule $Z$ satisfies $\textit{before}(i, j)$ if and only if $\textit{sol}(Z)$ satisfies $\textit{before}(i, j)$. The following result can then be proved easily by induction on the number of connectives in $\varphi$.

**Lemma 2** *For any PPE $\varphi$ over $\mathcal{A}$, if $s$ is a solution, then $\textit{sched}(s)$ is a valid schedule. If $Z$ is a valid schedule, then $\textit{sol}(Z)$ is a solution.*

The *minimum makespan* (for $\varphi$) is defined to be the infimum makespan over all valid schedules, i.e., the infimum of $\textit{make}(Z)$ over all valid schedules $Z$. The *minimum solution makespan* is defined to be the minimum makespan over all solutions, i.e., the minimum of $\textit{make}(s)$ over all solutions $s$. The following is the key result which links the schedule-based definition of minimum makespan with the solution-based definition. It follows from Proposition 5 and Lemma 2, since for any solution $s$ there is a valid schedule (i.e., $\textit{sched}(s)$) with the same value of makespan, and for any valid schedule $Z$ there is a solution (i.e., $\textit{sol}(Z)$) with at least as good a value of makespan.

**Proposition 6** *Let $\varphi$ be any positive precedence expression over $\mathcal{A}$. Then the minimum makespan for $\varphi$ is equal to the minimum solution makespan.*

### 7.1.4 Probabilistic Scheduling Problems based on PPEs

The probabilistic versions of the scheduling problems are defined in just the same way as for JSPs. The duration of each activity $A_i$ is now a random variable. A positive precedence expression is used to represent the constraints.

The further definitions in Sections 2 and 4 can all be immediately extended to this much more general setting. All the results of the paper still hold, *with exactly the same proofs*. In particular, with a probabilistic problem one associates a corresponding deterministic problem in just the same way; the lower bound results in Section 4.2 are based on the longest path characterization of makespan; the Monte Carlo approach (or at least its usefulness) relies on the fact that the makespan of a solution is equal to the makespan of the associated schedule. Furthermore, the algorithms in Section 5 extend, given that one has a method of solving the corresponding deterministic problem.

The ordering-based policies that we use (based on fixing a partial ordering of activities, irrespective of the sampled values of the durations) are known as *Earliest Start* policies (Radermacher, 1985). These and other policies have been studied for RCPSPs (see e.g., Stork, 2000, however the aim in that work is to minimize expected makespan, whereas we are attempting to minimize $\alpha$-makespan).

### 7.1.5 Different Optimization Functions

Because our approach for evaluating and comparing solutions is based on the use of Monte Carlo simulation to generate a sample distribution, our techniques are quite general.

Much of the work in the paper also generalizes immediately to other regular cost functions, where "regular" means that the function is monotonic in the sense that increasing the end of any activity in a schedule will not decrease the cost. A regular function based on any efficiently computable measurement of the sample distributions can be accommodated. For example, we could easily adapt to situations where the probability of extreme solutions is important by basing the optimization function on the maximum sampled makespan. Conversely, we could use measures of the tightness of the makespan distribution for situations where minimizing variance as a measure of the accuracy of a schedule is important. Furthermore, weighted combinations of such functions (e.g., the $\alpha$-makespan plus a measure of distribution tightness) could be easily incorporated.

We can also modify our approach to account for other ways of comparing solutions based on the sample distributions. For example, we could perform $t$-tests using the sample distributions to determine if one solution has a significantly lower expected makespan.

## 7.2 Toward Better Algorithms for Probabilistic JSPs

There are two directions for future work on the algorithms presented in this paper. First, B&B-N could be improved to make more use of deterministic techniques and/or to incorporate probabilistic reasoning into existing deterministic techniques. For example, a number of deterministic lower bound formulations for PERT networks exist in the operations research literature (Ludwig, Möhring, & Stork, 2001) that may be used to evaluate partial solutions. Similarly, perhaps the dominance rules presented by Daniels and Carrillo (1997) for the one-machine $\beta$-robustness problem can be generalized to multiple resources. Another approach to improving the B&B-N performance is to incorporate explicit reasoning about probability distributions into standard constraint propagation techniques. Techniques such as the longest path calculations and edge-finding make inferences based on the propagation of minimum and maximum values for temporal variables. We believe that many of these techniques can be adapted to reason about probabilistic intervals; this is related to work done, for example, on simple temporal networks with uncertainty (Morris, Muscettola, & Vidal, 2001; Tsamardinos, 2002).

A second direction for future work is the improvement of the heuristic algorithms. The key advantage of these algorithms is that they make use of deterministic techniques for scheduling: by transforming probabilistic problems into deterministic problems, we bring a significant set of existing tools to bear on the problem. Further developments of this approach include adaptively changing $q$-values during the search in order to find those that lead to solutions with better values of probabilistic makespan ($D_\alpha(s)$). A deeper understanding of the relationship between good deterministic solutions and good probabilistic solutions, building on the work here, is necessary to pursue this work in a principled fashion.

Of course, proactive techniques are not sufficient. In practice, schedules are dynamic and need to be adapted as new jobs arrive or existing jobs are canceled. At execution time, a reactive component is necessary to deal with unexpected (or sufficiently unlikely) disruptions that, nonetheless, can occur. A complete solution to scheduling under uncertainty needs to incorporate all these elements to reason about uncertainty at different levels of granularity and under different time pressures. See the work of Bidot, Vidal, Laborie and Beck (2007) for recent work in this direction.

## 7.3 Exploiting Unsound Lower Bounds in Constraint Programming

The B&B-DQ-L algorithm may represent a problem-solving approach that can be applied beyond the current application area. If we abstract away the probabilistic JSP application, the central idea of B&B-DQ-L is to exploit an unsound lower bound to (over)constrain the search and then to run subsequent searches with a gradually relaxed unsound lower bound. Such an approach may play to the strengths of constraint programming: searching within highly constrained spaces.

For example, the assignment problem (AP) is a well-known lower bound for the traveling salesman problem (TSP) and has been used as a cost-based constraint in the literature (Focacci, Lodi, & Milano, 2002; Rousseau, Gendreau, Pesant, & Focacci, 2004). Given a TSP, $P$, let $AP(P, q)$ be the corresponding assignment problem with the travel distances multiplied by $q$. That is, let $d_{ij}$ be the distance between cities $i$ and $j$ in $P$ and let $d'_{ij}$ be the distance between cities $i$ and $j$ in $AP(P, q)$. Then $d'_{ij} = d_{ij} \times q$ for $q \geq 1$. An approach similar to that of the B&B-DQ-L algorithm can now be applied to solve the TSP.

It would be interesting to investigate how the approach compares with the traditional optimization approach in constraint programming. It may be particularly useful in applications where the evaluation of partial solutions is very expensive but where there exists a parameterizable, inexpensive lower bound.

## 8. Conclusion

In this paper, we addressed job shop scheduling when the durations of the activities are independent random variables. A theoretical framework was created to formally define this problem and to prove the soundness of two algorithm components: Monte Carlo simulation to find upper bounds on the probabilistic makespan of a solution and a partial solution; and a carefully defined deterministic JSP whose optimal makespan is a lower bound on the probabilistic makespan of the corresponding probabilistic JSP.

We then used these two components together with either constraint programming or tabu search to define a number of algorithms to solve probabilistic JSPs. We introduced three solution approaches: a branch-and-bound technique using Monte Carlo simulation to evaluate partial solutions; an iterative deterministic search using Monte Carlo simulation to evaluate the solutions from a series of increasingly less constrained problems based on a parameterizable lower bound; and a number of deterministic filtering algorithms which generate a sequence of solutions to a deterministic JSP, each of which is then simulated using Monte Carlo simulation.

Our empirical evaluation demonstrated that the branch-and-bound technique is only able to find approximately optimal solutions for very small problem instances. The iterative deterministic search performs as well as, or better than, the branch-and-bound approach for all problem sizes. However, for medium and large instances, the deterministic filtering techniques perform much more strongly while providing no optimality guarantees. Further experimentation demonstrated that for the techniques using deterministic methods, the correlation between the deterministic makespan and probabilistic makespan is a key factor in algorithm performance: taking into account the variance of the duration in a deterministic problem led to strong correlations and good algorithmic performance.

Proactive scheduling techniques seek to incorporate models of uncertainty into an off-line, predictive schedule. The goal of such techniques is to increase the robustness of the schedules produced. This is important because a schedule is not typically generated or executed in isolation. Other decisions such as when to deliver raw materials and how to schedule up- and down-stream factories are all affected by an individual schedule. Indeed, a schedule can be seen as a locus of competing constraints from across a company and supply chain (Fox, 1983). Differences between a predictive schedule and its execution can be a significant source of disruption leading to cascading delays across widely separated entities. The ability, therefore, to develop schedules that are robust to uncertainty is very important. This paper represents a step in that direction.

## Acknowledgments

This work has received support from the Science Foundation Ireland under grants 00/PI.1/C075 and 05/IN/I886, the Natural Sciences and Engineering Research Council of Canada, and ILOG, SA. The authors would like to thank Daria Terekhov and Radoslaw Szymanek for comments on previous versions of the paper. Preliminary versions of the work reported in this paper have been published in Beck and Wilson (2004, 2005).